\documentclass{article}
\usepackage{ijcai18}

\usepackage{times}
\usepackage{xcolor}
\usepackage[utf8]{inputenc}
\usepackage[small]{caption}
\usepackage{amsmath,amsfonts,amssymb,amsthm,epsfig,epstopdf,url,array}

\usepackage{bbm}

\theoremstyle{plain}
\newtheorem{thm}{Theorem}[section]

\newtheorem{prop}[thm]{Proposition}

\theoremstyle{definition}
\newtheorem{defn}[thm]{Definition}
\newtheorem{exmp}{Example}[section]

\theoremstyle{remark}

\usepackage[utf8]{inputenc} 
\usepackage[T1]{fontenc}    
\usepackage{hyperref}       
\usepackage{url}            
\usepackage{booktabs}       
\usepackage{amsfonts}       
\usepackage{microtype}      

\usepackage{times}
\usepackage[english]{babel}
\usepackage{graphicx} 
\usepackage{subcaption}
\usepackage{tikz}
\usepackage{pgfplots}
\pgfplotsset{compat=newest}

\usepackage{natbib}
\setcitestyle{numbers}

\usepackage{algorithm}
\usepackage{algorithmic}
\usepackage{amsmath}

\usepackage{hyperref}

\newcommand{\tblref}[1]{Table~\ref{#1}}
\newcommand{\Tblref}[1]{Table~\ref{#1}}
\newcommand{\figref}[1]{Figure~\ref{#1}}
\newcommand{\Figref}[1]{Figure~\ref{#1}}
\newcommand{\secref}[1]{Section~\ref{#1}}

\newcommand{\best}[1]{{\fontseries{b}\selectfont #1}}

\newcommand{\domain}[1]{#1}
\newcommand{\sgn}{\text{sgn}}

\newskip\tblskipamount
\newcommand{\hlinetop}{\toprule}
\newcommand{\hlinemid}{\midrule}
\newcommand{\hlinebot}{\bottomrule}
\newcommand{\colsep}{2.1mm}
\newcommand{\leftsep}{0mm}
\newcommand{\subheadersep}{.4ex}
\newcommand{\methodsep}{.8ex}
\makeatletter
\newcommand{\tablesize}{\@setfontsize\small\@viiipt\@ixpt}
\newcommand{\tablestdsize}{\@setfontsize\small\@viipt\@ixpt}
\makeatother
\newcommand{\inlinestd}[1]{{\tablestdsize$\pm$#1}}
\newcommand{\withstd}[1]{\raisebox{0.5mm}{\tablestdsize$\pm$#1}}
\newcommand{\stddevs}{\vspace*{-0.5mm}}

\newcommand{\cp}{{RWA}}

\title{Unsupervised Domain Adaptation with Random Walks on Target Labelings}

\author{
Twan van Laarhoven,
Elena Marchiori
\\ 
Radboud University, Postbus 9010, 6500GL Nijmegen, The Netherlands \\
{tvanlaarhoven@cs.ru.nl},
{elenam@cs.ru.nl}
}

\begin{document}
\maketitle

\begin{abstract}
Unsupervised Domain Adaptation (DA) is used to automatize the task of labeling 
data: an unlabeled dataset (target) is annotated using a labeled dataset 
(source) from a related domain.
We cast domain adaptation as the problem of finding stable labels for target 
examples. A new definition of label stability is proposed, motivated by a generalization error bound for large margin linear classifiers: a target labeling is stable when, with 
high probability, a classifier trained on a random subsample of the target with 
that labeling yields the same labeling. We find stable labelings using a random 
walk on a directed graph with transition probabilities based on labeling 
stability. The majority vote of  those labelings visited by the walk yields a 
stable label for each target example.
The resulting domain adaptation algorithm is strikingly easy to implement and 
apply: It does not rely on data transformations, which are in general 
computational prohibitive in the presence of many input features, and does not 
need to access the source data, which is advantageous when data sharing is 
restricted. By acting on the original feature space, our method is able to take 
full advantage of deep features from external pre-trained neural networks, as 
demonstrated by the results of our experiments.
\end{abstract} 

\section{Introduction}
\label{intro}

Unsupervised domain adaptation (DA) addresses the problem of building a good predictor for a target domain using labeled training data from a related source domain and target unlabeled training data. A typical example in visual object recognition involves two different datasets  consisting of images taken under different cameras or conditions: for instance, one dataset consists of images taken at home  with a digital camera while another dataset contains images taken in a controlled environment  with  studio lightning conditions.

In some cases, the source domain is related to the target one,  but predictive features for the target domain may not even be present in the source domain as illustrated in the toy example in the figure. 
For instance this phenomenon can happen in natural language processing, where different genres often use very different vocabulary to described similar concepts. 
Here the target domain is rotated $45\deg$ compared to the source domain.
A linear classifier $h_s$ for the source domain will have an accuracy of only around 84\% on the target domain. If we perform feature selection on the source data, then we lose a feature that is relevant to the target domain and we will not be able to improve the accuracy.
However, the two classes are well separated in the target domain, and it should be possible to find a large-margin classifier separating the classes.

Just trying to separate the classes in the target domain is not enough, mainly because this does not tell us which class is which, since no labeled target data are available.
For that we need to use the relation to the source domain.
More generally, there is a trade-off between having a classifier that separates the classes in the target domain, and a classifier that stays close to the knowledge from the source domain.   We propose to model such trade-off by casting domain adaptation as the problem of finding a `stable'  label for each target example. 

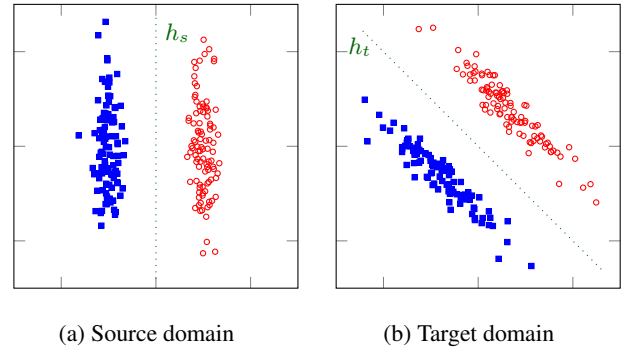
\begin{figure}\label{fig:toy1}
  \centering
  \begin{subfigure}{0.5\linewidth}
    \centering
    \begin{tikzpicture}
      \begin{axis}
        [width=1.25\linewidth
        ,height=1.25\linewidth
        ,enlargelimits=false,xmin=-3,xmax=3,ymin=-3,ymax=3
        ,xticklabel=\empty,yticklabel=\empty
        ,scatter/classes={0={mark=square*,blue,mark size=1},1={mark=o,red,mark size=1}}]
\addplot[scatter,only marks,scatter src=explicit symbolic] coordinates{
  (0.985781,-1.041493) [1]
  (-1.044921,0.864684) [0]
  (1.213682,0.073762) [1]
  (-0.933178,0.211031) [0]
  (1.144489,0.354293) [1]
  (-0.682819,-0.189106) [0]
  (0.970846,0.252990) [1]
  (-1.006211,-0.534247) [0]
  (1.087175,0.630152) [1]
  (-1.005279,1.071850) [0]
  (0.729782,-0.127590) [1]
  (-1.089346,0.218294) [0]
  (0.854970,-0.328184) [1]
  (-0.883802,-1.342938) [0]
  (1.164058,-0.705382) [1]
  (-0.819070,1.527633) [0]
  (1.148724,0.436236) [1]
  (-1.164185,-1.411341) [0]
  (1.091530,-1.288544) [1]
  (-0.871996,0.827334) [0]
  (0.936312,0.746649) [1]
  (-1.057615,-0.010865) [0]
  (0.879666,-0.266368) [1]
  (-0.896851,0.616293) [0]
  (1.263428,-1.069961) [1]
  (-1.072219,-0.390147) [0]
  (0.990290,-0.994329) [1]
  (-0.965438,0.594438) [0]
  (0.927881,-0.813409) [1]
  (-1.004581,0.680607) [0]
  (0.822512,1.331489) [1]
  (-1.195855,0.169966) [0]
  (1.028869,1.949459) [1]
  (-1.137605,-0.573040) [0]
  (1.084582,-0.438057) [1]
  (-0.749028,-1.043636) [0]
  (1.263061,0.772766) [1]
  (-1.066772,0.697427) [0]
  (0.810297,1.782085) [1]
  (-1.275362,-1.371011) [0]
  (0.901306,0.291008) [1]
  (-0.811347,0.073360) [0]
  (1.201868,-0.116083) [1]
  (-0.815831,-0.763550) [0]
  (1.251518,-2.229022) [1]
  (-0.794931,-0.352892) [0]
  (1.040500,-0.951500) [1]
  (-0.901525,-0.401071) [0]
  (0.898028,-0.098675) [1]
  (-0.943884,-0.585740) [0]
  (0.995169,-2.261561) [1]
  (-1.043800,-1.225521) [0]
  (1.091485,-0.784889) [1]
  (-1.100130,0.093703) [0]
  (0.857106,-0.733377) [1]
  (-1.241766,-0.570671) [0]
  (1.133046,-1.243604) [1]
  (-1.302507,-0.591844) [0]
  (0.987298,-0.856493) [1]
  (-0.991570,-0.020788) [0]
  (0.796297,0.482512) [1]
  (-1.032717,0.549557) [0]
  (1.062947,-1.489818) [1]
  (-1.016661,-0.466451) [0]
  (1.097550,0.752386) [1]
  (-0.983518,1.452086) [0]
  (1.124326,-1.173834) [1]
  (-1.128256,-0.929461) [0]
  (1.128155,0.129440) [1]
  (-1.027796,1.859736) [0]
  (0.689539,0.540751) [1]
  (-1.139425,-1.366602) [0]
  (0.982617,0.898425) [1]
  (-0.933357,-0.606442) [0]
  (0.830682,0.093793) [1]
  (-0.993241,0.938696) [0]
  (0.937062,-1.366754) [1]
  (-0.936145,-0.875002) [0]
  (0.876572,-0.007324) [1]
  (-0.703202,-0.891869) [0]
  (0.909184,-1.083129) [1]
  (-1.076836,1.365582) [0]
  (0.801306,0.126280) [1]
  (-0.890504,-0.671469) [0]
  (0.811026,1.479188) [1]
  (-1.132857,1.422761) [0]
  (0.838054,-0.187542) [1]
  (-0.884152,0.513115) [0]
  (0.939394,0.152264) [1]
  (-0.985732,-0.815464) [0]
  (1.000911,0.838712) [1]
  (-0.844022,0.212874) [0]
  (1.185136,-1.153208) [1]
  (-1.185165,0.337968) [0]
  (0.932846,-0.099331) [1]
  (-1.192357,-0.222446) [0]
  (0.695790,0.599267) [1]
  (-1.034431,-1.062659) [0]
  (0.832775,1.085605) [1]
  (-1.063681,-1.218980) [0]
  (1.339003,-0.358249) [1]
  (-0.904476,-1.146366) [0]
  (1.072053,0.801405) [1]
  (-0.959702,-1.188479) [0]
  (1.208721,-0.432247) [1]
  (-1.630885,0.227963) [0]
  (0.961467,-0.019941) [1]
  (-0.809297,0.044902) [0]
  (0.839598,1.656220) [1]
  (-1.023810,1.816745) [0]
  (0.865327,1.271358) [1]
  (-1.197919,-0.676539) [0]
  (1.041018,-0.533134) [1]
  (-1.115594,-0.159760) [0]
  (0.914771,-0.976076) [1]
  (-0.814744,-0.577822) [0]
  (0.924194,-0.874858) [1]
  (-0.871528,-0.543201) [0]
  (1.070209,0.946506) [1]
  (-1.222841,0.971803) [0]
  (1.222013,-0.480873) [1]
  (-1.295955,0.276528) [0]
  (1.075733,-2.021664) [1]
  (-1.053642,-0.997224) [0]
  (0.946082,-1.197251) [1]
  (-1.138916,0.235581) [0]
  (1.313724,0.245356) [1]
  (-0.912639,0.542492) [0]
  (0.991866,-0.062968) [1]
  (-1.079283,0.407020) [0]
  (1.068274,-0.377016) [1]
  (-1.125629,0.596237) [0]
  (0.751063,0.003984) [1]
  (-1.105837,0.349156) [0]
  (1.055774,0.118036) [1]
  (-1.069489,0.159010) [0]
  (1.217498,0.397683) [1]
  (-0.976918,-1.106315) [0]
  (0.949185,0.308867) [1]
  (-1.026210,-1.142506) [0]
  (0.829730,-1.305564) [1]
  (-1.217349,2.346184) [0]
  (0.999255,-0.260934) [1]
  (-0.975639,-0.176267) [0]
  (0.857427,0.946004) [1]
  (-1.097140,0.705622) [0]
  (1.105750,-0.499066) [1]
  (-0.719157,0.419001) [0]
  (0.921737,0.487246) [1]
  (-1.054066,-0.484327) [0]
  (0.998416,0.723755) [1]
  (-1.064958,2.630232) [0]
  (0.908205,-1.247471) [1]
  (-0.896690,-0.205944) [0]
  (1.171885,-0.304197) [1]
  (-1.152378,-1.678256) [0]
  (1.145233,0.475901) [1]
  (-1.234202,-0.940139) [0]
  (1.270390,-0.669940) [1]
  (-0.864420,-1.443307) [0]
  (1.114521,-0.322985) [1]
  (-0.964770,1.058205) [0]
  (1.064273,0.820173) [1]
  (-1.052502,0.205169) [0]
  (1.227004,1.798260) [1]
  (-1.098416,0.562038) [0]
  (1.090552,-0.051400) [1]
  (-0.982708,1.350704) [0]
  (0.941807,-0.325593) [1]
  (-0.800416,-0.202378) [0]
  (1.230981,1.850315) [1]
  (-0.742265,0.855192) [0]
  (0.661983,-0.223839) [1]
  (-0.806234,-0.778885) [0]
  (0.770777,-0.041616) [1]
  (-0.924149,0.504166) [0]
  (1.303853,-0.574182) [1]
  (-1.148628,-0.445025) [0]
  (0.887049,0.324794) [1]
  (-1.011707,1.257630) [0]
  (1.007922,2.252660) [1]
  (-1.172925,0.565135) [0]
  (0.820022,1.041063) [1]
  (-0.846821,-0.521067) [0]
  (0.925480,1.809837) [1]
  (-1.166040,-0.134341) [0]
  (0.723609,-0.752125) [1]
  (-1.050269,-0.791226) [0]
  (0.851357,0.698310) [1]
  (-1.335339,-0.459621) [0]
  (0.910589,-0.432338) [1]
  (-0.850108,-1.391470) [0]
  (0.924696,0.841831) [1]
  (-0.649184,0.107059) [0]
  (1.126098,2.051800) [1]
  (-1.140563,-0.115522) [0]
  (1.116436,-0.753720) [1]
  (-0.854898,0.810557) [0]
  (0.975788,-1.352491) [1]
  (-1.081992,-0.321225) [0]
};
        \draw[dotted,green!40!black] (axis cs: 0,-2.8) to (axis cs: 0,2.8) node [below right] {\small $h_s$};
      \end{axis}
    \end{tikzpicture}
    \caption{Source domain}
  \end{subfigure}%
  \begin{subfigure}{0.5\linewidth}
    \centering
    \begin{tikzpicture}
      \begin{axis}
        [width=1.25\linewidth
        ,height=1.25\linewidth
        ,enlargelimits=false,xmin=-3,xmax=3,ymin=-3,ymax=3
        ,xticklabel=\empty,yticklabel=\empty
        ,scatter/classes={0={mark=square*,blue,mark size=1},1={mark=o,red,mark size=1}}]
\addplot[scatter,only marks,scatter src=explicit symbolic] coordinates{
  (0.651091,0.407986) [1]
  (-0.378756,-1.212614) [0]
  (0.994658,0.536077) [1]
  (-1.341622,-0.656054) [0]
  (0.401290,1.067244) [1]
  (-0.608953,-1.185839) [0]
  (0.324315,0.907313) [1]
  (-1.607181,-0.120661) [0]
  (-0.075049,1.690070) [1]
  (-0.626344,-0.952775) [0]
  (0.824883,0.510549) [1]
  (-0.226334,-1.120140) [0]
  (0.558377,0.800743) [1]
  (-0.600422,-0.486838) [0]
  (0.452638,0.675632) [1]
  (-0.873927,-0.368071) [0]
  (0.520963,1.166572) [1]
  (-0.069354,-1.522132) [0]
  (1.248498,0.169189) [1]
  (-0.946757,-0.344863) [0]
  (0.198598,1.572467) [1]
  (-0.615681,-0.610381) [0]
  (0.891902,0.496841) [1]
  (0.616790,-2.025507) [0]
  (1.307105,0.104087) [1]
  (-2.338448,0.103472) [0]
  (0.457376,1.290881) [1]
  (-1.787521,0.497856) [0]
  (-0.096850,1.547112) [1]
  (-0.621908,-0.802068) [0]
  (2.373272,-0.801397) [1]
  (-0.995696,-0.571555) [0]
  (0.595690,0.460400) [1]
  (-0.209174,-1.073264) [0]
  (1.019053,0.858410) [1]
  (-0.596386,-1.002974) [0]
  (-0.071702,1.391390) [1]
  (-0.759101,-0.500675) [0]
  (1.433434,0.049790) [1]
  (0.256520,-1.512426) [0]
  (1.739247,-0.218887) [1]
  (-0.207343,-0.974034) [0]
  (-0.172690,1.517274) [1]
  (-0.179629,-1.132997) [0]
  (0.705587,0.614435) [1]
  (0.311090,-1.684494) [0]
  (1.257910,0.447240) [1]
  (-0.962106,-0.820868) [0]
  (0.541512,1.124364) [1]
  (-0.619346,-0.421241) [0]
  (-0.245482,1.764129) [1]
  (-0.559180,-0.909497) [0]
  (1.209260,0.418579) [1]
  (0.225842,-1.378294) [0]
  (0.241967,1.508071) [1]
  (-0.347367,-1.029012) [0]
  (1.089881,0.083587) [1]
  (-2.010842,0.396598) [0]
  (0.915048,0.506205) [1]
  (-1.861366,0.330881) [0]
  (0.164745,0.872197) [1]
  (-0.597061,-0.544626) [0]
  (0.144614,0.776721) [1]
  (-1.066769,-0.664709) [0]
  (0.955972,0.221444) [1]
  (-1.248124,-0.345223) [0]
  (1.120004,0.067215) [1]
  (0.622602,-1.585013) [0]
  (0.077056,1.535999) [1]
  (-0.771876,-0.397245) [0]
  (1.169125,0.111048) [1]
  (0.073608,-1.625009) [0]
  (1.536578,-0.122087) [1]
  (-1.508114,0.079759) [0]
  (0.751606,0.726006) [1]
  (-1.114158,-0.147062) [0]
  (0.107494,1.185002) [1]
  (-0.276928,-1.021663) [0]
  (0.318055,1.499428) [1]
  (-0.286278,-1.116420) [0]
  (-0.951905,2.506097) [1]
  (-1.229420,0.161663) [0]
  (0.576072,0.922214) [1]
  (-0.770543,-0.513381) [0]
  (1.709908,-0.796065) [1]
  (0.092669,-1.517449) [0]
  (1.540258,-0.119391) [1]
  (-0.582860,-1.341867) [0]
  (1.468058,0.074910) [1]
  (-1.295179,-0.158459) [0]
  (0.288654,1.301991) [1]
  (-0.523177,-0.962768) [0]
  (0.681222,0.915466) [1]
  (-1.343827,-0.177698) [0]
  (0.227422,1.154739) [1]
  (-0.080341,-1.324412) [0]
  (0.081939,1.501570) [1]
  (-1.244844,-0.074656) [0]
  (2.217024,-0.875180) [1]
  (-1.115084,-0.162436) [0]
  (0.909528,0.617530) [1]
  (-1.356184,-0.355713) [0]
  (-0.322468,1.520346) [1]
  (-1.209650,-0.102968) [0]
  (-0.453137,1.634905) [1]
  (-0.132374,-1.042124) [0]
  (2.493101,-1.187550) [1]
  (-0.724874,-0.461840) [0]
  (0.524382,1.023454) [1]
  (0.226411,-1.666855) [0]
  (1.327432,0.227683) [1]
  (0.430510,-2.378779) [0]
  (0.614434,1.110111) [1]
  (-1.606238,-0.017325) [0]
  (-0.225104,1.701386) [1]
  (-0.929754,-0.787789) [0]
  (0.652591,0.306757) [1]
  (-1.001082,-0.249826) [0]
  (0.257805,0.713687) [1]
  (-0.714815,-0.520249) [0]
  (1.513008,-0.001676) [1]
  (-0.126943,-1.431841) [0]
  (0.092064,0.874496) [1]
  (-0.548358,-0.984136) [0]
  (0.467710,0.688536) [1]
  (-0.944570,-0.417478) [0]
  (0.500882,1.154604) [1]
  (-0.757056,-0.734203) [0]
  (0.172557,1.176784) [1]
  (-1.447511,-0.004477) [0]
  (1.066789,-0.146008) [1]
  (-1.564311,0.379782) [0]
  (1.186896,0.216098) [1]
  (-0.452004,-1.014576) [0]
  (0.418782,0.851944) [1]
  (-1.601365,-0.043092) [0]
  (0.749300,0.564502) [1]
  (-1.479966,0.103760) [0]
  (0.185580,1.250597) [1]
  (-1.356297,-0.029119) [0]
  (1.346840,0.203594) [1]
  (-0.560053,-0.533950) [0]
  (1.140917,0.433434) [1]
  (-0.977234,-0.556763) [0]
  (0.403107,0.788612) [1]
  (-1.542590,0.009620) [0]
  (0.264536,1.110919) [1]
  (1.124695,-2.531123) [0]
  (0.440877,1.039307) [1]
  (-0.344082,-1.120078) [0]
  (0.352257,1.502272) [1]
  (-0.093282,-1.296580) [0]
  (0.977337,0.870690) [1]
  (-1.511976,-0.394928) [0]
  (0.615390,0.617453) [1]
  (-1.473573,0.161142) [0]
  (0.607470,1.181949) [1]
  (-1.274093,0.042657) [0]
  (0.193548,1.221842) [1]
  (-1.337837,-0.137885) [0]
  (-0.057378,1.577948) [1]
  (-1.368939,-0.097535) [0]
  (1.410178,-0.023559) [1]
  (-0.966020,-0.393136) [0]
  (0.368826,1.136256) [1]
  (-0.518309,-0.755529) [0]
  (-0.168546,1.385674) [1]
  (0.023966,-1.081240) [0]
  (1.921325,-0.020597) [1]
  (-1.217539,-0.306031) [0]
  (0.469558,0.999371) [1]
  (-1.497926,0.122902) [0]
  (0.706176,0.520261) [1]
  (-2.091291,0.664716) [0]
  (0.075083,1.196512) [1]
  (-1.005924,-0.306196) [0]
  (0.873780,0.263972) [1]
  (-1.047770,-0.183690) [0]
  (0.388556,1.131044) [1]
  (-0.812495,-0.639115) [0]
  (0.685452,0.854536) [1]
  (-1.083986,-0.645756) [0]
  (0.218934,1.044728) [1]
  (-0.741309,-0.294329) [0]
  (0.707783,0.616069) [1]
  (-0.069346,-1.559881) [0]
  (-0.042297,1.946432) [1]
  (-0.846230,-0.912708) [0]
  (-0.422036,2.034207) [1]
  (-0.568397,-0.970522) [0]
  (1.588873,-0.093320) [1]
  (-0.368028,-1.082744) [0]
  (1.672222,-0.135309) [1]
  (-0.869508,-0.441557) [0]
  (0.985671,0.498080) [1]
  (-0.664848,-0.814182) [0]
  (-1.255263,2.477487) [1]
  (-0.365634,-0.754706) [0]
  (2.047940,-0.263660) [1]
  (-2.394723,0.985045) [0]
};
        \draw[dotted,green!40!black] (axis cs: 2.6,-2.6) to (axis cs: -2.5,2.5) node [below] {\small $h_t$};
      \end{axis}
    \end{tikzpicture}
    \caption{Target domain}
  \end{subfigure}
  \caption{
    A simple dataset for DA, the vertical dimension is relevant for the target domain, but not for the source.
  }
\vskip -.1in
\end{figure}

We introduce the notion of labeling stability, motivated by a generalization bound for linear large margin classifiers: a target labeling is stable when, with high expectation, a target hypothesis trained on a random subsample of the target data with that labeling yields the same labeling.  
To find stable target labelings  we use a formalization based on random walks. We define a Markov chain with states equal to labelings from large margin linear classifiers and one-step transition probabilities defined using the proposed notion of labeling stability. 
Then we perform a random walk starting at the labeling obtained from the source hypothesis.  The walk will be attracted toward more stable labelings, which will be visited more often. The majority vote of the labelings visited by the walk provides our final estimated label for each target example.

We call the resulting unsupervised adaptation algorithm {\cp} ({\it R}andom {\it W}alk based {\it A}daptation). {\cp }  is strikingly simple to implement and apply. It does not rely on data transformations, which are in general computational prohibitive in the presence of many input features. {\cp}  does not need to access the  source data.  It acts on the original feature space, hence can take full advantage of the use of deep features from external pre-trained deep neural networks, as demonstrated by the results of our experiments. Results of extensive experiments on sentiment analysis and image object recognition show state-of-the-art performance of {\cp }  across adaptation datasets with diverse nature and characteristics. Notably, using deep learning features from pre-trained deep neural networks {\cp} outperforms much more involved end-to-end DA methods based on deep learning. 

Our contributions can be summarized as follows:  (1) a new definition of stability of a target labeling inspired by a generalization bound for linear large margin classifiers; (2) a new representation of the DA problem based on random walks; (3) a strikingly simple method for unsupervised DA;  (4) a direct and effective way to exploit deep features from pre-trained deep neural networks for visual adaptation tasks; (5) new state-of-the-art results on hard adaptation tasks with image as well as text data.

\section{Related work}\label{related}

The majority of algorithms for domain adaptation try to reduce the discrepancy between the source and target distributions using data transformation or augmentation, an approach motivated by theoretical studies. Many algorithms based on this approach have been proposed.
These algorithms mainly differ in the type of transformation used and in the underlying assumptions of the method. 
For instance, Subspace Alignment (SA)  \cite{Fernando2013SA} is a manifold based method that projects the source and target distributions into a lower-dimensional manifold and aligns the source
and target subspaces by computing a linear map that minimizes the Frobenius norm of their difference.
The more recent Correlation Alignment (CORAL) method \citep{sun2016return} performs domain adaptation by using a  transformation that minimizes the distance between the covariance of source and target.  These algorithms are the current state-of-the-art in the category of simple shallow methods for DA and will be used as baselines to assess the performance of our method.
A drawback of these methods is that they are not directly scalable to data with a high number of features. For instance, CORAL needs to compute and invert a covariance matrix. 

Methods based on self-training employ the source labeled data to train an initial model, which is then used to guess the labels of the target data. On the next round, the unlabeled data with pseudo labels are incorporated to train a new model. This procedure is iterated for a fixed number of times, or until convergence. Methods based on this approach, like  \cite{li2008self,bruzzone2010domain} differ in the way target samples are added and used. A drawback of these methods is their possible convergence to low-quality solutions when the source and target domain are not strongly related \cite{margolis2011literature}.

Recently end-to-end DA methods based on deep neural networks
have been shown to perform better than the  aforementioned approaches. However they need large train data \cite{sener2016learning}, use also a few labelled target examples to tune parameters \cite{long2016unsupervised} and are sensitive to (hyper-)parameters of the learning procedure \cite{ganin2016domain}.
 
Random walks have been used in many different contexts, in particular for semi-supervised learning, e.g.  \cite{szummer2002partially,zhou2005learning}. In these methods a similarity graph over labeled and unlabeled examples is used to perform label propagation. Therefore these works and our method differ in the graph representation as well as problem formulation. 

\section{Method}

Let $T$ be a set of unlabeled examples $x_t$ drawn from a target distribution ${\mathcal T}$ over the instance space $X$. Let $h_s$ be a source hypothesis trained on a set of labeled data with examples $x_s$ drawn from a source distribution ${\mathcal S}$. For simplicity, we focus on binary classification. So the source classifier is $\sgn h_s$. Nevertheless, the proposed method can be used with more than two classes, as shown in the experiments.  

As set {\cal H} of hypotheses we consider large margin linear Support Vector Machine (SVM) hypotheses $h_t$ over the target input space. Our set of candidate target labelings consists of target labelings induced by  hypotheses in {\cal H}.

The unsupervised domain adaptation problem is to find the true labels of examples in $T$ using $h_s$ and $T$. 
To solve this problem we propose to measure the quality of a labeling using  stability as described in the follow.

\subsection{Stability of a labeling}

Our definition of labeling stability is motivated by the following generalization bound for linear SVM classifiers.

\begin{thm}[Theorem 6.8 \cite{cristianini2000introduction}]
Consider thresholding real-value linear functions $f$ with unit weight vectors. For any probability distribution {\cal D} on $X\times \{-1,1\}$, with probability $1-\delta$ over $l$ random examples, the maximum margin hyperplane has error no more than 
\begin{equation}
{\it err}(g) \leq \frac{1}{l-d}(d \log\frac{el}{d} + \log\frac{l}{\delta}),
\end{equation}
where $d$ denotes the number of support vectors.
\end{thm}

The theorem shows that the fewer the number of support vectors the better generalization is expected.

Motivated by this result, we propose to cast DA as the problem of finding target labelings which yield SVM hypotheses with a small number of support vectors. To solve this problem, we relate the number of support vectors of an SVM to the stability of its predictions when trained on a random subsample of the dataset. Since the SVM hypothesis does not change when removing examples which are not support vectors, if there are few support vectors then the chance that at least one occurs in a random subsample will be small.

To exploit relatedness between source and target we  combine the source hypothesis $h_s$ with target hypotheses $h_t$ by simply averaging them $g=\frac{1}{2}(h_s+h_t)$. This choice amounts to consider $h_s$ and $h_t$ equally relevant.  Each average hypothesis $g$ can be mapped into a unique linear SVM hypothesis $h$ such that $\sgn(g(T)) = \sgn(h(T))$ by training the SVM on $(T,\sgn(g(T))$. Therefore, in the sequel, for such a $g$, we implicitly refer to its corresponding linear SVM $h$.

Now,  consider a candidate labeling $Y$. Let  $h$ be the SVM trained on $(T,Y)$ and $D$ the set of support vectors of $h$. For a  subsample $B$ of some size $m$ generated by random selection with replacement, the probability that $B$ contains all support vector of $h$ is 
\begin{equation*}
P(D\subseteq B) \ge \bigl(1 - (1 - 1/|T|)^m \bigr)^{|D|} \approx \bigl(1 - e^{-m/|T|} \bigr)^{|D|} .
\end{equation*}

If $D\subseteq B$, then an SVM trained on $B$ will be the same as $h$.
So, if $P(D\subseteq B)$ is large, then class predictions from an SVM trained on random subsamples of $(T,Y)$ and those from $h$ are likely to be the same. However, there may be many labelings satisfying this property. We need to consider the popularity of $Y$ also with respect to (SVM trained on) the other labelings. To do so, we propose the following formalization based on Markov chains.  

Consider the Markov chain with single-step transition probabilities $p_{ij}$  of jumping from any node (state) $i$ to one of its adjacent nodes $j$  defined as

$$p_{ij} : = P_{B} [\sgn((g_B)(T)) = Y^j],$$ 

where $B$ is a random subsample of $(T,Y^i)$ of size $m$, and $g_B = h_s+h_B$, with $h_B$ the linear SVM hypothesis trained on $B$. By the above argument, a hypothesis with few support vectors is more robust under the transitions $p_{ij}$, that is, $p_{ii}$ is higher.
In particular, it is not hard to prove the following inequality.
\begin{prop}
  $p_{ii} \ge P(D_i \subseteq B)$,
where $D_i$ is the set of support vectors of a SVM trained on $(T,Y^i)$.
\end{prop}
Consider Markov chain $p_{11}=0.5$, $p_{12}=0.5$, $p_{13}=0$, $p_{21}=0$, $p_{22}=0.1$, $p_{23}= 0.9$,  $p_{31}=0$ $p_{32}=0.5$, $p_{33}=0.5$. Although states $1$ and $3$ have both maximum $p_{ii}$, state $3$ is more popular and should be preferred. Popularity can be measured by the stationary distribution $\pi$. For a Markov chain $P = [p_{ij}]$, $\pi$ is a solution to the equations 
\begin{equation*}
\pi P = \pi.
\end{equation*}\label{eq:stability}
The above equations expanded say that for every $i$, $\sum_j \pi(j) p_{ji} = \pi(i)$, that is, executing one step of the chain starting with the distribution $\pi$ results in the same distribution.

If the directed graph over candidate labelings with weights $p_{ij}$ is strongly connected then, by the  fundamental theorem of Markov chains, $\pi$ is unique.   Under this assumption we can define the stability of labeling $Y^i$ as 
\begin{defn}[stability of a labeling]
$s_m(Y^i) := \pi(i)$.
\end{defn}
$\pi(i)$ is inversely proportional to the expected amount of time to return to state $i$ given that the walk started in state $i$.  So $Y$ with a high degree of stability will be obtained by many hypotheses trained on subsamples of $T$ equipped with a labeling $Y$. 

Computing the stationary distribution of our Markov process is in general computationally hard, because the number of our labelings is $|T|^{O(d)}$, where $d$ is the dimension of the feature space, and the computation of $p_{ij}$ has complexity exponential in the number of random subsamples.   we can sample target labelings from this distribution

\begin{figure*}
  \centering
  \def\sz{4cm}
  \begin{subfigure}{0.3\linewidth}
    \centering
    \begin{tikzpicture}
      \begin{axis}
        [width=\sz
        ,height=\sz
        ,enlargelimits=false,xmin=-5,xmax=5,ymin=-5,ymax=5
        ,xticklabel=\empty,yticklabel=\empty
        ,scatter/classes={-1={mark=square*,blue,mark size=1},1={mark=o,red,mark size=1}}]
\addplot[scatter,only marks,scatter src=explicit symbolic] coordinates{
  (3.107302,0.061812) [1]
  (-1.963980,-0.668468) [-1]
  (3.355827,-0.770869) [1]
  (-3.427289,0.000552) [-1]
  (3.459749,-0.042477) [1]
  (-3.237791,-0.342877) [-1]
  (2.766284,1.219701) [1]
  (-3.062890,-0.399841) [-1]
  (1.888177,1.882706) [1]
  (-3.121537,-2.422418) [-1]
  (3.204679,-0.297598) [1]
  (-2.791990,-0.518602) [-1]
  (3.198165,1.458789) [1]
  (-2.525399,0.994962) [-1]
  (2.985614,0.211262) [1]
  (-1.845585,-1.956906) [-1]
  (3.710753,3.251299) [1]
  (-3.162177,-1.356084) [-1]
  (3.919251,-0.503494) [1]
  (-2.581425,-0.497151) [-1]
  (3.505163,-1.159508) [1]
  (-3.092968,-0.502955) [-1]
  (2.954180,0.754256) [1]
  (-3.547936,-0.184158) [-1]
  (2.802327,-0.300813) [1]
  (-2.595753,1.848840) [-1]
  (3.137377,-1.842463) [1]
  (-2.387212,2.213973) [-1]
  (3.184989,-0.228032) [1]
  (-3.222646,0.330624) [-1]
  (2.545325,-1.095048) [1]
  (-3.166310,0.207253) [-1]
  (2.868342,-0.595292) [1]
  (-4.242751,0.951554) [-1]
  (2.353956,-0.840725) [1]
  (-3.296552,-1.477961) [-1]
  (3.548465,0.909868) [1]
  (-2.757312,0.546579) [-1]
  (4.049550,0.978052) [1]
  (-2.392048,1.589064) [-1]
  (2.468231,-0.304568) [1]
  (-3.049203,-0.232453) [-1]
  (3.026991,0.381611) [1]
  (-3.548999,1.108461) [-1]
  (3.299192,0.314749) [1]
  (-2.806928,0.018862) [-1]
  (2.347187,-0.766171) [1]
  (-2.939882,-0.809877) [-1]
  (4.201537,-1.440621) [1]
  (-3.072768,-0.303192) [-1]
  (3.167534,-1.385445) [1]
  (-2.571755,-2.141980) [-1]
  (3.846622,-1.314045) [1]
  (-3.795689,0.327992) [-1]
  (3.348362,-2.506369) [1]
  (-3.010783,-0.502158) [-1]
  (2.831313,1.630907) [1]
  (-3.123778,0.954331) [-1]
  (4.057106,0.625469) [1]
  (-3.670154,0.031661) [-1]
  (2.720345,0.573173) [1]
  (-3.194171,0.434538) [-1]
  (2.096194,-0.876187) [1]
  (-3.370662,-0.851232) [-1]
  (2.931299,0.678298) [1]
  (-2.796727,0.519343) [-1]
  (2.274087,-0.460943) [1]
  (-2.682443,0.492686) [-1]
  (2.435811,-1.446532) [1]
  (-3.996117,-0.195982) [-1]
  (3.618480,0.317423) [1]
  (-2.821345,-0.567673) [-1]
  (3.412637,-0.481641) [1]
  (-2.953172,1.290351) [-1]
  (3.625041,0.449470) [1]
  (-2.502902,-1.529440) [-1]
  (2.827401,1.807418) [1]
  (-3.321975,-0.081491) [-1]
  (2.061209,0.603636) [1]
  (-2.483442,-0.181289) [-1]
  (3.204908,-0.454643) [1]
  (-3.466500,0.128413) [-1]
  (2.569035,0.801278) [1]
  (-3.075730,0.314923) [-1]
  (3.114288,-1.859437) [1]
  (-2.720431,-0.062056) [-1]
  (2.990261,0.516764) [1]
  (-2.833018,1.276591) [-1]
  (2.228111,-0.194543) [1]
  (-2.733291,0.231572) [-1]
  (2.891493,-0.424126) [1]
  (-3.555329,-1.752407) [-1]
  (2.280399,1.314353) [1]
  (-3.596335,1.656289) [-1]
  (2.764085,1.227358) [1]
  (-3.152483,-1.200907) [-1]
  (2.784624,0.098619) [1]
  (-3.227053,-0.446981) [-1]
  (3.564601,0.036213) [1]
  (-2.571908,0.175932) [-1]
  (3.174328,0.117947) [1]
  (-3.524359,-0.594214) [-1]
  (2.473899,-0.660989) [1]
  (-2.966121,0.525832) [-1]
  (3.091259,0.809425) [1]
  (-3.290404,0.075295) [-1]
  (2.888501,-0.194226) [1]
  (-3.343353,0.398639) [-1]
  (2.942138,-0.477856) [1]
  (-2.637295,-1.459479) [-1]
  (2.658942,0.052107) [1]
  (-3.447235,1.417886) [-1]
  (2.506507,-0.251667) [1]
  (-2.462241,0.215860) [-1]
  (2.086524,0.124910) [1]
  (-2.806753,-1.289110) [-1]
  (1.960354,-0.160758) [1]
  (-3.370236,0.116926) [-1]
  (2.878472,1.351235) [1]
  (-2.646658,-0.724087) [-1]
  (3.560450,0.812746) [1]
  (-3.268364,-0.491211) [-1]
  (3.328036,0.875991) [1]
  (-2.642832,0.840650) [-1]
  (2.764706,-1.655455) [1]
  (-2.490922,0.322578) [-1]
  (3.086989,0.430880) [1]
  (-2.323813,-0.168253) [-1]
  (1.947388,-0.744519) [1]
  (-1.957233,0.280319) [-1]
  (2.719424,-0.413663) [1]
  (-3.334948,-0.530179) [-1]
  (2.770777,1.184250) [1]
  (-2.622241,-1.007143) [-1]
  (3.927614,0.264678) [1]
  (-2.539371,-0.633033) [-1]
  (2.696279,2.271897) [1]
  (-3.664407,-0.429497) [-1]
  (2.728970,0.105553) [1]
  (-2.712067,1.100529) [-1]
  (3.258058,1.620745) [1]
  (-2.669547,0.423476) [-1]
  (2.376917,-0.616390) [1]
  (-3.104523,-0.941165) [-1]
  (2.510429,-0.668659) [1]
  (-3.994726,0.679311) [-1]
  (2.767668,-0.824844) [1]
  (-2.473271,-0.178266) [-1]
  (3.001999,1.005759) [1]
  (-1.921460,0.804222) [-1]
  (3.924931,0.290273) [1]
  (-2.637985,-2.213924) [-1]
  (2.541575,0.721866) [1]
  (-3.572253,-1.688612) [-1]
  (2.763344,-0.024487) [1]
  (-3.666799,-0.578752) [-1]
  (3.527780,-1.467015) [1]
  (-3.156939,-0.955918) [-1]
  (3.219202,-1.170857) [1]
  (-2.842313,0.624901) [-1]
  (3.030389,-0.056708) [1]
  (-2.901873,-2.114683) [-1]
  (2.383781,0.164259) [1]
  (-3.998976,0.805930) [-1]
  (3.053563,0.942052) [1]
  (-3.056783,0.823023) [-1]
  (3.227512,2.312675) [1]
  (-3.173368,1.957758) [-1]
  (3.360756,2.226741) [1]
  (-2.637237,-1.032206) [-1]
  (3.541565,-0.082730) [1]
  (-2.409355,1.784791) [-1]
  (3.290885,-0.363689) [1]
  (-3.538975,-1.308390) [-1]
  (3.063531,0.661735) [1]
  (-3.397526,-1.920116) [-1]
  (2.704022,-0.885643) [1]
  (-3.006413,0.524712) [-1]
  (3.450976,-1.050130) [1]
  (-3.310489,1.304774) [-1]
  (3.314911,-1.583786) [1]
  (-3.122530,0.738658) [-1]
  (3.305866,0.372805) [1]
  (-2.491644,0.401092) [-1]
  (1.720128,1.150750) [1]
  (-2.947108,1.037209) [-1]
  (2.491961,-0.804888) [1]
  (-3.479259,1.272266) [-1]
  (1.742275,-1.015261) [1]
  (-2.843086,-0.106025) [-1]
  (2.519121,1.148025) [1]
  (-3.443018,-0.666663) [-1]
  (2.292032,0.372744) [1]
  (-3.281803,0.205062) [-1]
  (2.953165,-0.710178) [1]
  (-2.896829,-1.450402) [-1]
  (2.629278,-0.722046) [1]
  (-2.883507,0.376207) [-1]
  (2.707481,1.118178) [1]
  (-3.569517,-0.217475) [-1]
};
        \draw[dotted,green!40!black] (axis cs: 0,-4.6) to (axis cs: 0,4.6) node [below right] {\small $h_s$};
      \end{axis}
    \end{tikzpicture}
    \caption{Source data and classifier.}
  \end{subfigure}%
  \begin{subfigure}{0.3\linewidth}
    \centering
    \begin{tikzpicture}
      \begin{axis}
        [width=\sz
        ,height=\sz
        ,enlargelimits=false,xmin=-5,xmax=5,ymin=-5,ymax=5
        ,xticklabel=\empty,yticklabel=\empty
        ,scatter/classes={-1={mark=square*,blue,mark size=1},1={mark=o,red,mark size=1}}]
\addplot[scatter,only marks,scatter src=explicit symbolic] coordinates{
  (1.527195,3.207841) [1]
  (-0.318188,-2.093072) [-1]
  (-0.000151,3.371984) [1]
  (0.549367,-2.633527) [-1]
  (1.333580,2.784308) [1]
  (0.165983,-2.907534) [-1]
  (0.525929,3.311786) [1]
  (-0.505895,-2.025744) [-1]
  (1.094870,3.332329) [1]
  (-0.934411,-2.677923) [-1]
  (-1.624815,3.034535) [1]
  (0.805578,-3.183573) [-1]
  (1.180630,2.522243) [1]
  (-0.988504,-2.862551) [-1]
  (1.357128,2.697678) [1]
  (0.491379,-2.320740) [-1]
  (0.873905,2.696769) [1]
  (-0.589201,-3.551753) [-1]
  (-0.047068,3.332536) [1]
  (-1.078235,-3.324117) [-1]
  (1.566932,3.178296) [1]
  (0.473813,-2.870010) [-1]
  (0.287218,2.871393) [1]
  (-0.222835,-2.921086) [-1]
  (2.957835,2.924005) [1]
  (0.598683,-2.944457) [-1]
  (0.669108,2.498379) [1]
  (-0.818702,-3.154090) [-1]
  (0.763861,3.207379) [1]
  (0.716946,-3.373174) [-1]
  (-2.250985,2.931831) [1]
  (-1.779224,-2.620255) [-1]
  (1.663342,3.201731) [1]
  (-0.541714,-3.619850) [-1]
  (1.300521,3.047832) [1]
  (-0.977470,-2.341950) [-1]
  (1.671678,2.896224) [1]
  (2.562190,-3.012654) [-1]
  (-0.325264,3.037170) [1]
  (-0.092090,-2.720273) [-1]
  (0.605800,3.562646) [1]
  (0.915060,-2.020170) [-1]
  (-0.025841,2.067541) [1]
  (-3.939953,-2.990718) [-1]
  (-1.952309,2.807983) [1]
  (-0.557568,-3.888405) [-1]
  (-0.136830,2.773228) [1]
  (-1.595695,-3.568973) [-1]
  (-0.880632,2.816696) [1]
  (0.842755,-3.273496) [-1]
  (0.621080,2.858309) [1]
  (0.943874,-1.602389) [-1]
  (-1.097535,2.426337) [1]
  (-0.664589,-2.334399) [-1]
  (-0.578133,2.217010) [1]
  (0.208410,-2.671659) [-1]
  (0.196451,3.118399) [1]
  (1.381016,-3.169612) [-1]
  (-0.446541,1.695894) [1]
  (-1.676186,-3.156661) [-1]
  (-1.078245,2.581404) [1]
  (-1.354291,-3.006855) [-1]
  (1.880484,3.711372) [1]
  (-0.452649,-4.362556) [-1]
  (-0.305473,2.311686) [1]
  (0.344927,-2.313092) [-1]
  (0.277904,3.428992) [1]
  (-2.021209,-3.124094) [-1]
  (0.462718,2.071980) [1]
  (-1.041420,-3.719477) [-1]
  (0.670350,2.773844) [1]
  (-0.537625,-2.647563) [-1]
  (0.049699,2.874193) [1]
  (0.163698,-3.057678) [-1]
  (-0.744385,2.272807) [1]
  (-2.440971,-2.194731) [-1]
  (1.106741,2.781861) [1]
  (0.593707,-3.216557) [-1]
  (0.803409,3.111730) [1]
  (-0.419338,-3.293364) [-1]
  (0.149558,2.508918) [1]
  (0.192731,-2.572763) [-1]
  (1.332729,3.881461) [1]
  (-0.779171,-4.141592) [-1]
  (0.203775,2.779215) [1]
  (-1.073690,-3.201827) [-1]
  (0.673538,2.679607) [1]
  (-0.883532,-3.831854) [-1]
  (1.011328,3.308015) [1]
  (0.835970,-2.775868) [-1]
  (0.101624,3.529783) [1]
  (-2.915453,-3.273070) [-1]
  (-0.307351,3.221061) [1]
  (-1.816090,-2.403804) [-1]
  (2.725270,2.801064) [1]
  (0.138356,-2.549537) [-1]
  (1.246466,2.552455) [1]
  (1.162322,-3.800717) [-1]
  (-2.004223,3.244211) [1]
  (-0.759656,-2.907141) [-1]
  (0.950159,2.914661) [1]
  (-0.377369,-3.260380) [-1]
  (0.623204,2.210587) [1]
  (0.517538,-2.753579) [-1]
  (1.299892,2.386942) [1]
  (-0.677471,-3.146854) [-1]
  (0.333312,2.997774) [1]
  (1.263133,-3.359586) [-1]
  (-0.335675,3.232361) [1]
  (-1.147816,-2.523327) [-1]
  (0.663933,2.844839) [1]
  (-0.007109,-3.973520) [-1]
  (2.329148,3.409790) [1]
  (0.124654,-3.288038) [-1]
  (1.366459,2.415347) [1]
  (-1.118745,-2.642970) [-1]
  (0.101988,2.965097) [1]
  (0.485199,-1.868249) [-1]
  (-0.253628,3.200833) [1]
  (-0.823080,-2.414581) [-1]
  (0.232768,2.615722) [1]
  (0.293087,-3.186778) [-1]
  (0.531463,2.910918) [1]
  (1.265692,-3.399721) [-1]
  (0.145305,2.982079) [1]
  (-0.656640,-2.643762) [-1]
  (0.080543,2.811947) [1]
  (-0.356787,-2.905281) [-1]
  (0.369001,3.289990) [1]
  (-1.113783,-3.184305) [-1]
  (1.608464,1.978809) [1]
  (0.624277,-3.068515) [-1]
  (1.278732,3.020545) [1]
  (0.358891,-3.106165) [-1]
  (0.793418,2.586441) [1]
  (0.305404,-4.167647) [-1]
  (-0.200014,3.226901) [1]
  (-1.548800,-2.569594) [-1]
  (1.129990,3.237471) [1]
  (0.030647,-3.349241) [-1]
  (0.079510,2.635842) [1]
  (0.481258,-3.561649) [-1]
  (1.941381,2.330069) [1]
  (1.071451,-3.695981) [-1]
  (1.439769,3.498662) [1]
  (-2.163306,-2.626742) [-1]
  (0.861017,3.613110) [1]
  (-1.895925,-2.698495) [-1]
  (0.808234,2.781086) [1]
  (1.815419,-3.754420) [-1]
  (0.653039,2.696052) [1]
  (-0.590679,-3.348813) [-1]
  (0.783932,3.526454) [1]
  (-0.687406,-4.182976) [-1]
  (-0.038092,3.244984) [1]
  (-1.006264,-2.970125) [-1]
  (0.627753,3.046274) [1]
  (-0.253118,-3.001138) [-1]
  (-0.083516,2.569780) [1]
  (-0.475374,-2.492108) [-1]
  (-0.124591,3.128378) [1]
  (-0.295830,-2.927769) [-1]
  (0.372078,3.420355) [1]
  (0.776039,-3.870983) [-1]
  (0.995526,3.272761) [1]
  (1.484293,-2.724328) [-1]
  (1.786200,2.560747) [1]
  (0.088268,-3.963316) [-1]
  (0.416616,2.153230) [1]
  (-0.415814,-3.564222) [-1]
  (0.390615,2.971033) [1]
  (0.020551,-2.898401) [-1]
  (1.036385,2.740629) [1]
  (-1.556914,-2.898719) [-1]
  (0.068743,2.732476) [1]
  (1.171001,-3.169875) [-1]
  (0.055013,2.985846) [1]
  (-0.490986,-3.516908) [-1]
  (-0.778644,3.206189) [1]
  (-2.438794,-3.036450) [-1]
  (-1.077344,2.546084) [1]
  (1.488678,-2.896522) [-1]
  (1.127737,2.827249) [1]
  (-2.527760,-2.913991) [-1]
  (0.923800,3.294668) [1]
  (0.710431,-2.709238) [-1]
  (-1.203904,3.029265) [1]
  (0.666966,-3.621774) [-1]
  (-0.449308,2.562624) [1]
  (-0.649148,-2.967461) [-1]
  (-0.490632,3.026273) [1]
  (0.055704,-3.159359) [-1]
  (-0.437123,2.725004) [1]
  (-1.498003,-2.304785) [-1]
  (0.387077,4.061791) [1]
  (-0.734699,-2.081537) [-1]
  (-0.530169,3.063788) [1]
  (-1.408754,-3.582843) [-1]
  (1.396600,3.172348) [1]
  (0.767338,-3.488410) [-1]
};
        \draw[dotted,green!40!black] (axis cs: 4.6,-0.6) to (axis cs: -4.5,0.5) node [above right] {\small $h_B$};
      \end{axis}
    \end{tikzpicture}
    \caption{Target set and $h_B$ of the last iteration.}
  \end{subfigure}
  \begin{subfigure}{0.3\linewidth}
    \centering
    \begin{tikzpicture}
      \begin{axis}
        [width=\sz
        ,height=\sz
        ,enlargelimits=false,xmin=-5,xmax=5,ymin=-5,ymax=5
        ,xticklabel=\empty,yticklabel=\empty
        ,scatter/classes={-1={mark=square*,blue,mark size=1},1={mark=o,red,mark size=1}}]
\addplot[scatter,only marks,scatter src=explicit symbolic] coordinates{
  (3.107302,0.061812) [1]
  (-1.963980,-0.668468) [-1]
  (3.355827,-0.770869) [1]
  (-3.427289,0.000552) [-1]
  (3.459749,-0.042477) [1]
  (-3.237791,-0.342877) [-1]
  (2.766284,1.219701) [1]
  (-3.062890,-0.399841) [-1]
  (1.888177,1.882706) [1]
  (-3.121537,-2.422418) [-1]
  (3.204679,-0.297598) [1]
  (-2.791990,-0.518602) [-1]
  (3.198165,1.458789) [1]
  (-2.525399,0.994962) [-1]
  (2.985614,0.211262) [1]
  (-1.845585,-1.956906) [-1]
  (3.710753,3.251299) [1]
  (-3.162177,-1.356084) [-1]
  (3.919251,-0.503494) [1]
  (-2.581425,-0.497151) [-1]
  (3.505163,-1.159508) [1]
  (-3.092968,-0.502955) [-1]
  (2.954180,0.754256) [1]
  (-3.547936,-0.184158) [-1]
  (2.802327,-0.300813) [1]
  (-2.595753,1.848840) [-1]
  (3.137377,-1.842463) [1]
  (-2.387212,2.213973) [-1]
  (3.184989,-0.228032) [1]
  (-3.222646,0.330624) [-1]
  (2.545325,-1.095048) [1]
  (-3.166310,0.207253) [-1]
  (2.868342,-0.595292) [1]
  (-4.242751,0.951554) [-1]
  (2.353956,-0.840725) [1]
  (-3.296552,-1.477961) [-1]
  (3.548465,0.909868) [1]
  (-2.757312,0.546579) [-1]
  (4.049550,0.978052) [1]
  (-2.392048,1.589064) [-1]
  (2.468231,-0.304568) [1]
  (-3.049203,-0.232453) [-1]
  (3.026991,0.381611) [1]
  (-3.548999,1.108461) [-1]
  (3.299192,0.314749) [1]
  (-2.806928,0.018862) [-1]
  (2.347187,-0.766171) [1]
  (-2.939882,-0.809877) [-1]
  (4.201537,-1.440621) [1]
  (-3.072768,-0.303192) [-1]
  (3.167534,-1.385445) [1]
  (-2.571755,-2.141980) [-1]
  (3.846622,-1.314045) [1]
  (-3.795689,0.327992) [-1]
  (3.348362,-2.506369) [1]
  (-3.010783,-0.502158) [-1]
  (2.831313,1.630907) [1]
  (-3.123778,0.954331) [-1]
  (4.057106,0.625469) [1]
  (-3.670154,0.031661) [-1]
  (2.720345,0.573173) [1]
  (-3.194171,0.434538) [-1]
  (2.096194,-0.876187) [1]
  (-3.370662,-0.851232) [-1]
  (2.931299,0.678298) [1]
  (-2.796727,0.519343) [-1]
  (2.274087,-0.460943) [1]
  (-2.682443,0.492686) [-1]
  (2.435811,-1.446532) [1]
  (-3.996117,-0.195982) [-1]
  (3.618480,0.317423) [1]
  (-2.821345,-0.567673) [-1]
  (3.412637,-0.481641) [1]
  (-2.953172,1.290351) [-1]
  (3.625041,0.449470) [1]
  (-2.502902,-1.529440) [-1]
  (2.827401,1.807418) [1]
  (-3.321975,-0.081491) [-1]
  (2.061209,0.603636) [1]
  (-2.483442,-0.181289) [-1]
  (3.204908,-0.454643) [1]
  (-3.466500,0.128413) [-1]
  (2.569035,0.801278) [1]
  (-3.075730,0.314923) [-1]
  (3.114288,-1.859437) [1]
  (-2.720431,-0.062056) [-1]
  (2.990261,0.516764) [1]
  (-2.833018,1.276591) [-1]
  (2.228111,-0.194543) [1]
  (-2.733291,0.231572) [-1]
  (2.891493,-0.424126) [1]
  (-3.555329,-1.752407) [-1]
  (2.280399,1.314353) [1]
  (-3.596335,1.656289) [-1]
  (2.764085,1.227358) [1]
  (-3.152483,-1.200907) [-1]
  (2.784624,0.098619) [1]
  (-3.227053,-0.446981) [-1]
  (3.564601,0.036213) [1]
  (-2.571908,0.175932) [-1]
  (3.174328,0.117947) [1]
  (-3.524359,-0.594214) [-1]
  (2.473899,-0.660989) [1]
  (-2.966121,0.525832) [-1]
  (3.091259,0.809425) [1]
  (-3.290404,0.075295) [-1]
  (2.888501,-0.194226) [1]
  (-3.343353,0.398639) [-1]
  (2.942138,-0.477856) [1]
  (-2.637295,-1.459479) [-1]
  (2.658942,0.052107) [1]
  (-3.447235,1.417886) [-1]
  (2.506507,-0.251667) [1]
  (-2.462241,0.215860) [-1]
  (2.086524,0.124910) [1]
  (-2.806753,-1.289110) [-1]
  (1.960354,-0.160758) [1]
  (-3.370236,0.116926) [-1]
  (2.878472,1.351235) [1]
  (-2.646658,-0.724087) [-1]
  (3.560450,0.812746) [1]
  (-3.268364,-0.491211) [-1]
  (3.328036,0.875991) [1]
  (-2.642832,0.840650) [-1]
  (2.764706,-1.655455) [1]
  (-2.490922,0.322578) [-1]
  (3.086989,0.430880) [1]
  (-2.323813,-0.168253) [-1]
  (1.947388,-0.744519) [1]
  (-1.957233,0.280319) [-1]
  (2.719424,-0.413663) [1]
  (-3.334948,-0.530179) [-1]
  (2.770777,1.184250) [1]
  (-2.622241,-1.007143) [-1]
  (3.927614,0.264678) [1]
  (-2.539371,-0.633033) [-1]
  (2.696279,2.271897) [1]
  (-3.664407,-0.429497) [-1]
  (2.728970,0.105553) [1]
  (-2.712067,1.100529) [-1]
  (3.258058,1.620745) [1]
  (-2.669547,0.423476) [-1]
  (2.376917,-0.616390) [1]
  (-3.104523,-0.941165) [-1]
  (2.510429,-0.668659) [1]
  (-3.994726,0.679311) [-1]
  (2.767668,-0.824844) [1]
  (-2.473271,-0.178266) [-1]
  (3.001999,1.005759) [1]
  (-1.921460,0.804222) [-1]
  (3.924931,0.290273) [1]
  (-2.637985,-2.213924) [-1]
  (2.541575,0.721866) [1]
  (-3.572253,-1.688612) [-1]
  (2.763344,-0.024487) [1]
  (-3.666799,-0.578752) [-1]
  (3.527780,-1.467015) [1]
  (-3.156939,-0.955918) [-1]
  (3.219202,-1.170857) [1]
  (-2.842313,0.624901) [-1]
  (3.030389,-0.056708) [1]
  (-2.901873,-2.114683) [-1]
  (2.383781,0.164259) [1]
  (-3.998976,0.805930) [-1]
  (3.053563,0.942052) [1]
  (-3.056783,0.823023) [-1]
  (3.227512,2.312675) [1]
  (-3.173368,1.957758) [-1]
  (3.360756,2.226741) [1]
  (-2.637237,-1.032206) [-1]
  (3.541565,-0.082730) [1]
  (-2.409355,1.784791) [-1]
  (3.290885,-0.363689) [1]
  (-3.538975,-1.308390) [-1]
  (3.063531,0.661735) [1]
  (-3.397526,-1.920116) [-1]
  (2.704022,-0.885643) [1]
  (-3.006413,0.524712) [-1]
  (3.450976,-1.050130) [1]
  (-3.310489,1.304774) [-1]
  (3.314911,-1.583786) [1]
  (-3.122530,0.738658) [-1]
  (3.305866,0.372805) [1]
  (-2.491644,0.401092) [-1]
  (1.720128,1.150750) [1]
  (-2.947108,1.037209) [-1]
  (2.491961,-0.804888) [1]
  (-3.479259,1.272266) [-1]
  (1.742275,-1.015261) [1]
  (-2.843086,-0.106025) [-1]
  (2.519121,1.148025) [1]
  (-3.443018,-0.666663) [-1]
  (2.292032,0.372744) [1]
  (-3.281803,0.205062) [-1]
  (2.953165,-0.710178) [1]
  (-2.896829,-1.450402) [-1]
  (2.629278,-0.722046) [1]
  (-2.883507,0.376207) [-1]
  (2.707481,1.118178) [1]
  (-3.569517,-0.217475) [-1]
};
\addplot[scatter,only marks,scatter src=explicit symbolic] coordinates{
  (1.527195,3.207841) [1]
  (-0.318188,-2.093072) [-1]
  (-0.000151,3.371984) [1]
  (0.549367,-2.633527) [-1]
  (1.333580,2.784308) [1]
  (0.165983,-2.907534) [-1]
  (0.525929,3.311786) [1]
  (-0.505895,-2.025744) [-1]
  (1.094870,3.332329) [1]
  (-0.934411,-2.677923) [-1]
  (-1.624815,3.034535) [1]
  (0.805578,-3.183573) [-1]
  (1.180630,2.522243) [1]
  (-0.988504,-2.862551) [-1]
  (1.357128,2.697678) [1]
  (0.491379,-2.320740) [-1]
  (0.873905,2.696769) [1]
  (-0.589201,-3.551753) [-1]
  (-0.047068,3.332536) [1]
  (-1.078235,-3.324117) [-1]
  (1.566932,3.178296) [1]
  (0.473813,-2.870010) [-1]
  (0.287218,2.871393) [1]
  (-0.222835,-2.921086) [-1]
  (2.957835,2.924005) [1]
  (0.598683,-2.944457) [-1]
  (0.669108,2.498379) [1]
  (-0.818702,-3.154090) [-1]
  (0.763861,3.207379) [1]
  (0.716946,-3.373174) [-1]
  (-2.250985,2.931831) [1]
  (-1.779224,-2.620255) [-1]
  (1.663342,3.201731) [1]
  (-0.541714,-3.619850) [-1]
  (1.300521,3.047832) [1]
  (-0.977470,-2.341950) [-1]
  (1.671678,2.896224) [1]
  (2.562190,-3.012654) [-1]
  (-0.325264,3.037170) [1]
  (-0.092090,-2.720273) [-1]
  (0.605800,3.562646) [1]
  (0.915060,-2.020170) [-1]
  (-0.025841,2.067541) [1]
  (-3.939953,-2.990718) [-1]
  (-1.952309,2.807983) [1]
  (-0.557568,-3.888405) [-1]
  (-0.136830,2.773228) [1]
  (-1.595695,-3.568973) [-1]
  (-0.880632,2.816696) [1]
  (0.842755,-3.273496) [-1]
  (0.621080,2.858309) [1]
  (0.943874,-1.602389) [-1]
  (-1.097535,2.426337) [1]
  (-0.664589,-2.334399) [-1]
  (-0.578133,2.217010) [1]
  (0.208410,-2.671659) [-1]
  (0.196451,3.118399) [1]
  (1.381016,-3.169612) [-1]
  (-0.446541,1.695894) [1]
  (-1.676186,-3.156661) [-1]
  (-1.078245,2.581404) [1]
  (-1.354291,-3.006855) [-1]
  (1.880484,3.711372) [1]
  (-0.452649,-4.362556) [-1]
  (-0.305473,2.311686) [1]
  (0.344927,-2.313092) [-1]
  (0.277904,3.428992) [1]
  (-2.021209,-3.124094) [-1]
  (0.462718,2.071980) [1]
  (-1.041420,-3.719477) [-1]
  (0.670350,2.773844) [1]
  (-0.537625,-2.647563) [-1]
  (0.049699,2.874193) [1]
  (0.163698,-3.057678) [-1]
  (-0.744385,2.272807) [1]
  (-2.440971,-2.194731) [-1]
  (1.106741,2.781861) [1]
  (0.593707,-3.216557) [-1]
  (0.803409,3.111730) [1]
  (-0.419338,-3.293364) [-1]
  (0.149558,2.508918) [1]
  (0.192731,-2.572763) [-1]
  (1.332729,3.881461) [1]
  (-0.779171,-4.141592) [-1]
  (0.203775,2.779215) [1]
  (-1.073690,-3.201827) [-1]
  (0.673538,2.679607) [1]
  (-0.883532,-3.831854) [-1]
  (1.011328,3.308015) [1]
  (0.835970,-2.775868) [-1]
  (0.101624,3.529783) [1]
  (-2.915453,-3.273070) [-1]
  (-0.307351,3.221061) [1]
  (-1.816090,-2.403804) [-1]
  (2.725270,2.801064) [1]
  (0.138356,-2.549537) [-1]
  (1.246466,2.552455) [1]
  (1.162322,-3.800717) [-1]
  (-2.004223,3.244211) [1]
  (-0.759656,-2.907141) [-1]
  (0.950159,2.914661) [1]
  (-0.377369,-3.260380) [-1]
  (0.623204,2.210587) [1]
  (0.517538,-2.753579) [-1]
  (1.299892,2.386942) [1]
  (-0.677471,-3.146854) [-1]
  (0.333312,2.997774) [1]
  (1.263133,-3.359586) [-1]
  (-0.335675,3.232361) [1]
  (-1.147816,-2.523327) [-1]
  (0.663933,2.844839) [1]
  (-0.007109,-3.973520) [-1]
  (2.329148,3.409790) [1]
  (0.124654,-3.288038) [-1]
  (1.366459,2.415347) [1]
  (-1.118745,-2.642970) [-1]
  (0.101988,2.965097) [1]
  (0.485199,-1.868249) [-1]
  (-0.253628,3.200833) [1]
  (-0.823080,-2.414581) [-1]
  (0.232768,2.615722) [1]
  (0.293087,-3.186778) [-1]
  (0.531463,2.910918) [1]
  (1.265692,-3.399721) [-1]
  (0.145305,2.982079) [1]
  (-0.656640,-2.643762) [-1]
  (0.080543,2.811947) [1]
  (-0.356787,-2.905281) [-1]
  (0.369001,3.289990) [1]
  (-1.113783,-3.184305) [-1]
  (1.608464,1.978809) [1]
  (0.624277,-3.068515) [-1]
  (1.278732,3.020545) [1]
  (0.358891,-3.106165) [-1]
  (0.793418,2.586441) [1]
  (0.305404,-4.167647) [-1]
  (-0.200014,3.226901) [1]
  (-1.548800,-2.569594) [-1]
  (1.129990,3.237471) [1]
  (0.030647,-3.349241) [-1]
  (0.079510,2.635842) [1]
  (0.481258,-3.561649) [-1]
  (1.941381,2.330069) [1]
  (1.071451,-3.695981) [-1]
  (1.439769,3.498662) [1]
  (-2.163306,-2.626742) [-1]
  (0.861017,3.613110) [1]
  (-1.895925,-2.698495) [-1]
  (0.808234,2.781086) [1]
  (1.815419,-3.754420) [-1]
  (0.653039,2.696052) [1]
  (-0.590679,-3.348813) [-1]
  (0.783932,3.526454) [1]
  (-0.687406,-4.182976) [-1]
  (-0.038092,3.244984) [1]
  (-1.006264,-2.970125) [-1]
  (0.627753,3.046274) [1]
  (-0.253118,-3.001138) [-1]
  (-0.083516,2.569780) [1]
  (-0.475374,-2.492108) [-1]
  (-0.124591,3.128378) [1]
  (-0.295830,-2.927769) [-1]
  (0.372078,3.420355) [1]
  (0.776039,-3.870983) [-1]
  (0.995526,3.272761) [1]
  (1.484293,-2.724328) [-1]
  (1.786200,2.560747) [1]
  (0.088268,-3.963316) [-1]
  (0.416616,2.153230) [1]
  (-0.415814,-3.564222) [-1]
  (0.390615,2.971033) [1]
  (0.020551,-2.898401) [-1]
  (1.036385,2.740629) [1]
  (-1.556914,-2.898719) [-1]
  (0.068743,2.732476) [1]
  (1.171001,-3.169875) [-1]
  (0.055013,2.985846) [1]
  (-0.490986,-3.516908) [-1]
  (-0.778644,3.206189) [1]
  (-2.438794,-3.036450) [-1]
  (-1.077344,2.546084) [1]
  (1.488678,-2.896522) [-1]
  (1.127737,2.827249) [1]
  (-2.527760,-2.913991) [-1]
  (0.923800,3.294668) [1]
  (0.710431,-2.709238) [-1]
  (-1.203904,3.029265) [1]
  (0.666966,-3.621774) [-1]
  (-0.449308,2.562624) [1]
  (-0.649148,-2.967461) [-1]
  (-0.490632,3.026273) [1]
  (0.055704,-3.159359) [-1]
  (-0.437123,2.725004) [1]
  (-1.498003,-2.304785) [-1]
  (0.387077,4.061791) [1]
  (-0.734699,-2.081537) [-1]
  (-0.530169,3.063788) [1]
  (-1.408754,-3.582843) [-1]
  (1.396600,3.172348) [1]
  (0.767338,-3.488410) [-1]
};
        \draw[dotted,green!40!black] (axis cs: 4.4,-4.6) to (axis cs: -4.4,4.6) node [right] {\small $h$};
      \end{axis}
    \end{tikzpicture}
    \caption{Final classifier $h_s+h_B$.}
  \end{subfigure}
  \begin{subfigure}{0.3\linewidth}
    \begin{tikzpicture}
      \begin{axis}
        [width=\sz
        ,height=\sz
        ,xmin=0, xmax=13
        ,ymin=55, ymax=100
        ,xlabel={Iteration}, ylabel={Accuracy (\%)}]
\addplot+[] coordinates{
  (0.000000,65.000000)
  (1.000000,93.000000)
  (2.000000,99.500000)
  (3.000000,100.000000)
  (4.000000,100.000000)
  (5.000000,100.000000)
  (6.000000,100.000000)
  (7.000000,100.000000)
  (8.000000,100.000000)
  (9.000000,100.000000)
  (10.000000,100.000000)
  (11.000000,100.000000)
  (12.000000,100.000000)
  (13.000000,100.000000)
  (14.000000,100.000000)
  (15.000000,100.000000)
  (16.000000,100.000000)
  (17.000000,100.000000)
  (18.000000,100.000000)
  (19.000000,100.000000)
  (20.000000,100.000000)
  (21.000000,100.000000)
  (22.000000,100.000000)
  (23.000000,100.000000)
  (24.000000,100.000000)
  (25.000000,100.000000)
  (26.000000,100.000000)
  (27.000000,100.000000)
  (28.000000,100.000000)
  (29.000000,100.000000)
  (30.000000,100.000000)
  (31.000000,100.000000)
  (32.000000,100.000000)
  (33.000000,100.000000)
  (34.000000,100.000000)
  (35.000000,100.000000)
  (36.000000,100.000000)
  (37.000000,100.000000)
  (38.000000,100.000000)
  (39.000000,100.000000)
  (40.000000,100.000000)
  (41.000000,100.000000)
  (42.000000,100.000000)
  (43.000000,100.000000)
  (44.000000,100.000000)
  (45.000000,100.000000)
  (46.000000,100.000000)
  (47.000000,100.000000)
  (48.000000,100.000000)
  (49.000000,100.000000)
  (50.000000,100.000000)
  (51.000000,100.000000)
  (52.000000,100.000000)
  (53.000000,100.000000)
  (54.000000,100.000000)
  (55.000000,100.000000)
  (56.000000,100.000000)
  (57.000000,100.000000)
  (58.000000,100.000000)
  (59.000000,100.000000)
  (60.000000,100.000000)
  (61.000000,100.000000)
  (62.000000,100.000000)
  (63.000000,100.000000)
  (64.000000,100.000000)
  (65.000000,100.000000)
  (66.000000,100.000000)
  (67.000000,100.000000)
  (68.000000,100.000000)
  (69.000000,100.000000)
  (70.000000,100.000000)
  (71.000000,100.000000)
  (72.000000,100.000000)
  (73.000000,100.000000)
  (74.000000,100.000000)
  (75.000000,100.000000)
  (76.000000,100.000000)
  (77.000000,100.000000)
  (78.000000,100.000000)
  (79.000000,100.000000)
  (80.000000,100.000000)
  (81.000000,100.000000)
  (82.000000,100.000000)
  (83.000000,100.000000)
  (84.000000,100.000000)
  (85.000000,100.000000)
  (86.000000,100.000000)
  (87.000000,100.000000)
  (88.000000,100.000000)
  (89.000000,100.000000)
  (90.000000,100.000000)
  (91.000000,100.000000)
  (92.000000,100.000000)
  (93.000000,100.000000)
  (94.000000,100.000000)
  (95.000000,100.000000)
  (96.000000,100.000000)
  (97.000000,100.000000)
  (98.000000,100.000000)
  (99.000000,100.000000)
  (100.000000,100.000000)
};
      \end{axis}
    \end{tikzpicture}
    \caption{Accuracy on target domain}
  \end{subfigure}
  \begin{subfigure}{0.3\linewidth}
    \begin{tikzpicture}
      \begin{axis}
        [width=\sz
        ,height=\sz
        ,xmin=0, xmax=13
        ,ymin=0, ymax=0.2
        ,xlabel={Iteration}, ylabel={Margin}]
\addplot+[] coordinates{
  (0.000000,0.004447)
  (1.000000,0.024267)
  (2.000000,0.073257)
  (3.000000,0.196708)
  (4.000000,0.196708)
  (5.000000,0.196708)
  (6.000000,0.196708)
  (7.000000,0.196708)
  (8.000000,0.196708)
  (9.000000,0.196708)
  (10.000000,0.196708)
  (11.000000,0.196708)
  (12.000000,0.196708)
  (13.000000,0.196708)
  (14.000000,0.196708)
  (15.000000,0.196708)
  (16.000000,0.196708)
  (17.000000,0.196708)
  (18.000000,0.196708)
  (19.000000,0.196708)
  (20.000000,0.196708)
  (21.000000,0.196708)
  (22.000000,0.196708)
  (23.000000,0.196708)
  (24.000000,0.196708)
  (25.000000,0.196708)
  (26.000000,0.196708)
  (27.000000,0.196708)
  (28.000000,0.196708)
  (29.000000,0.196708)
  (30.000000,0.196708)
  (31.000000,0.196708)
  (32.000000,0.196708)
  (33.000000,0.196708)
  (34.000000,0.196708)
  (35.000000,0.196708)
  (36.000000,0.196708)
  (37.000000,0.196708)
  (38.000000,0.196708)
  (39.000000,0.196708)
  (40.000000,0.196708)
  (41.000000,0.196708)
  (42.000000,0.196708)
  (43.000000,0.196708)
  (44.000000,0.196708)
  (45.000000,0.196708)
  (46.000000,0.196708)
  (47.000000,0.196708)
  (48.000000,0.196708)
  (49.000000,0.196708)
  (50.000000,0.196708)
  (51.000000,0.196708)
  (52.000000,0.196708)
  (53.000000,0.196708)
  (54.000000,0.196708)
  (55.000000,0.196708)
  (56.000000,0.196708)
  (57.000000,0.196708)
  (58.000000,0.196708)
  (59.000000,0.196708)
  (60.000000,0.196708)
  (61.000000,0.196708)
  (62.000000,0.196708)
  (63.000000,0.196708)
  (64.000000,0.196708)
  (65.000000,0.196708)
  (66.000000,0.196708)
  (67.000000,0.196708)
  (68.000000,0.196708)
  (69.000000,0.196708)
  (70.000000,0.196708)
  (71.000000,0.196708)
  (72.000000,0.196708)
  (73.000000,0.196708)
  (74.000000,0.196708)
  (75.000000,0.196708)
  (76.000000,0.196708)
  (77.000000,0.196708)
  (78.000000,0.196708)
  (79.000000,0.196708)
  (80.000000,0.196708)
  (81.000000,0.196708)
  (82.000000,0.196708)
  (83.000000,0.196708)
  (84.000000,0.196708)
  (85.000000,0.196708)
  (86.000000,0.196708)
  (87.000000,0.196708)
  (88.000000,0.196708)
  (89.000000,0.196708)
  (90.000000,0.196708)
  (91.000000,0.196708)
  (92.000000,0.196708)
  (93.000000,0.196708)
  (94.000000,0.196708)
  (95.000000,0.196708)
  (96.000000,0.196708)
  (97.000000,0.196708)
  (98.000000,0.196708)
  (99.000000,0.196708)
  (100.000000,0.196708)
};
      \end{axis}
    \end{tikzpicture}
    \caption{Margin}
  \end{subfigure}
  \caption{
    A run of {\cp } on a simple dataset for domain adaptation.
  }
  \label{fig:toy1}
\end{figure*}
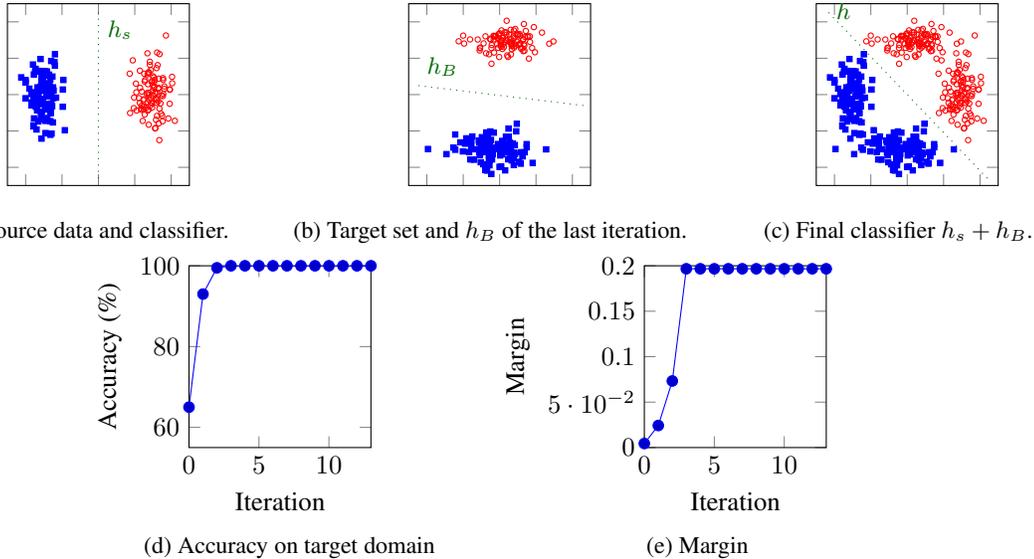

\subsection{Random walk over labelings}

Therefore our practical solution is to draw samples from the stationary distribution by running the Markov chain for sufficiently long. Then we choose as the final label of each target example the one assigned most frequently by the nodes (labelings) visited by the walk (the majority vote choice). 
If  $S(Y^i)$ is large, then $Y^i$ has higher a probability of being visited by a walk. Hence target labelings with a higher degree of stability are more likely to be selected. 

The resulting DA algorithm  is shown in Algorithm~\ref{alg1}. {\cp } considers class-balanced bootstrap samples, in order to rule out trivial solutions which assign the same label to all examples. Class balance has been shown to be important in semi-supervised learning, e.g. \cite{chapelle2008optimization} and  transductive learning, e.g. \cite{collobert2006large}. In our setting, this choice amounts to assume a uniform prior over the target class distribution. The following toy example illustrates our method.

\begin{exmp}
Consider source labeled dataset $S = [(-8,-1),(8,1)]$ and  target unlabeled dataset $T = [-9,-1,1,+9]$. The chosen values of source and target examples are not crucial.
There are three feasible target labelings:  $Y_1=(-1,-1,-1,+1)$, $Y_2=(-1,-1,+1,+1)$, $Y_3=(-1,+1,+1,+1)$.  Using balanced bootstrap samples of size $4$, we get the Markov chain shown in the diagram, where state $i$ denotes labeling $Y^i$. 
\usetikzlibrary{calc, automata, chains, arrows}
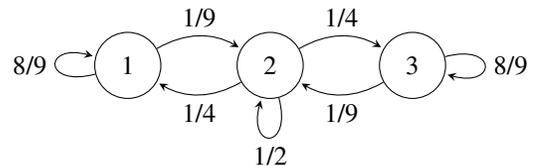
\begin{center}
\begin{tikzpicture}[
    start chain = going right,shorten >=1pt,auto,
    >=stealth, arr/.style={->}
    ]
  \foreach \i in {1,...,3} 
    \node[state, on chain]  (\i) {\i};
  \path[arr]  (1) edge[bend left] node {1/9} (2);
  \path[arr]  (2) edge[bend left] node {1/4} (1);
  \path[arr]  (2) edge[bend left] node {1/4} (3);
  \path[arr]  (3) edge[bend left] node {1/9} (2);
  \path[arr]  (1) edge[loop left] node {8/9} (1);
  \path[arr]  (2) edge[loop below] node {1/2} (2);
  \path[arr]  (3) edge[loop right] node {8/9} (3);
\end{tikzpicture}
\end{center}
This Markov chain is irreducible with stationary distribution $\pi(1)=\pi(3)=9/22$; $\pi(2)=4/22$. 
Therefore when starting from target labeling $Y_2$ from the source classifier $h_s(x) = sign(x)$ we will repeatedly reach each state.  A sufficiently long walk will contain all three labelings.  The majority vote label of target examples $-9$ and $+9$ is $-1$ and $+1$, since all labelings agree on that. The label of example $-1$ will be $-1$ because this is the `vote' from all occurrences of $Y_1$ and $Y_2$, which are expected to exceed those of $Y_3$.  Analogously, the label of example $+1$ will be $+1$.
\end{exmp}

%



%
%


The following toy example illustrates the execution of {\cp } on an artificial dataset.
\begin{exmp}
Figure \ref{fig:toy1} shows a typical run of {\cp }, with $K=15$ and $m=n=100$, on an artificial dataset: \ref{fig:toy1}.a) shows the source hypothesis $h_s$ and the original source data. Note that $h_s$ only is used by {\cp }, not the source data; \ref{fig:toy1}.b) plots the target hypothesis $h_B$ generated at the last iteration of {\cp }, and \ref{fig:toy1}.c) shows the combined hypothesis $h_s+h_B$, which makes source and target closer, resulting in a good hypothesis for both of them.
\end{exmp}

\begin{algorithm}[tb]
  \caption{\cp}
  \label{alg1}
  \begin{algorithmic}
    \STATE $Y^0 = \sgn(h_s(T))$
    \FOR{$k=1$ {\bfseries to} $K$}
      \STATE $B_k =$ class-balanced bootstrap sample from $(T,Y^{k-1})$
\STATE  $h^k = h_{B_k}+ h_s$
\STATE $Y^k = \sgn(h^k(T))$ 
    \ENDFOR
    \STATE  $Y = \text{MajorityVote}(Y^1, \dots, Y^K)$
  \end{algorithmic}
\end{algorithm}



\section{Experimental analysis}

We test the performance of {\cp } comparatively on $28$ adaptation tasks.
%
%
We assess all considered algorithms in a fully transductive setup where all unlabeled instances of the target are used during training for predicting their labels.
No labeled target domain data is used.
We evaluate the accuracy as the fraction of correctly labeled target instances. 

We compare with two state of the art shallow methods: Correlation Alignment (CORAL) \citep{sun2016return},
and Subspace Alignment (SA) \citep{Fernando2013SA}. 
Furthermore, on the Office 31 dataset, we compare also with published results reported in \cite{long2016deep}{ (based on ResNet (50 layers) features or architecture \citep{he2016deep}) of the following end-to-end deep learning methods for domain adaptation.
Deep Domain Confusion (DDC) \citep{tzeng2014deep},
Deep Adaptation Network (DAN) \citep{long2015learning},
Residual Transfer Network (RTN) \citep{long2016unsupervised},
Reverse Gradient (RevGrad)  \citep{ganin2015unsupervised} and
Joint Adaptation Networks (JAN) \citep{long2016deep}.
%
We run experiments with source code of the shallow DA methods. For deep DA methods we report the accuracies under the same evaluation protocol from the corresponding papers.  

In our experiments {\cp} uses labeled source instances only to train the source hypothesis, with internal cross-validation to select the regularization parameter.  For efficiency reasons we consider the same regularization parameter value for all provisional target hypotheses, computed by internal cross-validation on the target dataset labeled using the source hypothesis.
In all experiments we perform $K=500$ iterations. 
The bootstrap samples have the same size as the target dataset but are class balanced, that is, the size of each class in a bootstrap sample is equal to the  number of elements in the target set divided by the  number of classes. We investigate the sensitivity of these choices in \secref{sec:sensitivity}.
 

We use the linear SVM implemented by liblinear \citep{Fan2008liblinear}. For multi-class tasks we use a one-versus-all strategy. 
The SVM parameter $C$ is fixed to the same value for all target hypotheses. Such value is  obtained by tuning $C$ using $T$ with labels from $h_s$.


\subsection{Results}
We perform extensive experiments with text and image benchmark datasets of diverse characteristics: data with high number of features, data with a relatively small sample size, data with a larger number of classes and large scale data.

\subsubsection*{Amazon sentiment dataset}
This dataset \cite{blitzer2007biographies}  involves $4$ domains \domain{Books} (B), \domain{Dvd} (D), \domain{Electronics} (E) and \domain{Kitchen} (K), each with 1000 positive and 1000 negative examples obtained from the dichotomized 5-star rating.
There are over 400000 features, which are word unigram and bigram counts.
The number of features is too large for most domain adaptation methods. Therefore \citet{gong2013connecting} used feature selection to reduce the data set to 400 features. We conduct experiments with this reduced feature set and validation protocol as in \cite{sun2016return}: random subsamples of the source (1600 samples) and target (400 samples) data and standardized features. The experiment is repeated 20 times. 

We also report results of {\cp} under the same experimental protocol but with {\it all} features.
In this case we can not standardize the data, because that would destroy the sparsity, instead we normalize by dividing each feature by its standard deviation.
\Tblref{tbl:accuracy-amazon2-subset} reports average accuracy, which shows that, contrary to reports in previous works, using all features is beneficial. 

\begin{table}
 \caption{Accuracy on the Amazon sentiment dataset using the standard protocol of \citet{gong2013connecting,sun2016return}. Mean and standard deviation over 20 runs are shown.}
 \label{tbl:accuracy-amazon2-subset}
 \tablesize
 \centering
\begin{tabular}{@{\hspace*{\leftsep}}ll@{\hspace*{\colsep}}c@{\hspace*{\colsep}}c@{\hspace*{\colsep}}c@{\hspace*{\colsep}}c@{\hspace*{\colsep}}c}
\hlinetop
& & K$\to$D & D$\to$B & B$\to$E & E$\to$K & avg\\
\hlinemid
\multicolumn{5}{l}{400 features}\\[\subheadersep]
& Source SVM & {73.3}\inlinestd{1.9} & {78.3}\inlinestd{2.3} & {75.6}\inlinestd{1.5} & {83.1}\inlinestd{1.8} & {77.6}\inlinestd{1.1}\\[\methodsep]
& SA & {73.3}\inlinestd{1.9} & {78.3}\inlinestd{2.3} & {75.6}\inlinestd{1.5} & {83.1}\inlinestd{1.8} & {77.6}\inlinestd{1.1}\\
& CORAL & {73.5}\inlinestd{1.8} & {78.3}\inlinestd{2.0} & {76.1}\inlinestd{1.7} & {83.1}\inlinestd{1.9} & {77.7}\inlinestd{0.8}\\[\methodsep]
& RWA & {74.5}\inlinestd{2.2} & {78.8}\inlinestd{2.2} & {78.0}\inlinestd{2.0} & {83.8}\inlinestd{2.1} & {78.8}\inlinestd{1.0}\\
\hlinemid
\multicolumn{5}{l}{All features}\\[\subheadersep]
& Source SVM & {73.8}\inlinestd{2.3} & {78.8}\inlinestd{2.1} & {72.6}\inlinestd{2.3} & {85.9}\inlinestd{1.7} & {77.8}\inlinestd{1.1}\\[\methodsep]
& RWA & \best{76.7}\inlinestd{2.8} & \best{80.9}\inlinestd{2.2} & \best{78.8}\inlinestd{2.7} & \best{86.2}\inlinestd{2.2} & \best{80.7}\inlinestd{1.4}\\
\hlinebot
\end{tabular}
\end{table}

\subsubsection*{Office-Caltech 10 object recognition dataset}
This dataset \citep{gong2012geodesic} consists of $10$ classes of images from an office environment in 4 image domains: Webcam (W), DSLR (D), Amazon (A), and Caltech256 (C).
The dataset uses 800 SURF features, which we preprocess by dividing by the instance-wise mean followed by standardizing.
We follow the standard protocol \citep{gong2012geodesic,Fernando2013SA,sun2016return}, and use 20 labeled samples per class from the source domain (except for the DSLR source domain, for which we use 8 samples per class).
We repeated this experiment 20 times, and report the average accuracy in \tblref{tbl:accuracy-office-caltech-standard}. 
 {\cp}   achieves state-of-the-art results, with a substantial increase in accuracy over no adaptation (from 39\% to 47\% for {C$\to$W}).

In addition to the SURF features, we construct deep features from the Resnet network (50 layers)  \citep{he2016deep}, a deep neural network that was pre-trained on Imagenet dataset.
We rescale the images to $288\times288$ pixels, and then take 9 different crops of $224\times224$ pixels which we pass through the network. We then repeat this procedure for the horizontally flipped image, and we use the output of the nonlinearities on the last hidden layer as features, averaging over the different crops and flips. This corresponds roughly to the common data augmentation strategy used when the network was trained.
With these deep features, {\cp} shows a substantial increase in the accuracy compared to no adaptation  and to other adaptation methods.

\begin{table*}[t]
 \caption{Accuracy on the Office-Caltech 10 dataset, using the standard protocol of \citet{gong2012geodesic,Fernando2013SA,sun2016return}. Mean and standard deviation over 20 runs are shown.}
 \label{tbl:accuracy-office-caltech-standard}
 \tablesize
 \centering
\begin{tabular}{@{\hspace*{\leftsep}}ll@{\hspace*{\colsep}}c@{\hspace*{\colsep}}c@{\hspace*{\colsep}}c@{\hspace*{\colsep}}c@{\hspace*{\colsep}}c@{\hspace*{\colsep}}c@{\hspace*{\colsep}}c@{\hspace*{\colsep}}c@{\hspace*{\colsep}}c@{\hspace*{\colsep}}c@{\hspace*{\colsep}}c@{\hspace*{\colsep}}c@{\hspace*{\colsep}}c}
\hlinetop
& & A$\to$C & A$\to$D & A$\to$W & C$\to$A & C$\to$D & C$\to$W & D$\to$A & D$\to$C & D$\to$W & W$\to$A & W$\to$C & W$\to$D & avg\\
\hlinemid
\multicolumn{13}{l}{SURF features}\\[\subheadersep]
& Source SVM & {36.3} & {36.7} & {35.9} & {45.0} & {42.1} & {39.1} & {34.6} & {32.1} & {75.8} & {37.9} & {33.9} & {73.5} & {43.6}\\
\stddevs&&\withstd{1.6}&\withstd{3.3}&\withstd{2.5}&\withstd{2.2}&\withstd{3.3}&\withstd{3.8}&\withstd{1.0}&\withstd{0.9}&\withstd{2.4}&\withstd{0.5}&\withstd{1.2}&\withstd{3.4}&\withstd{1.0}\\[\methodsep]
& SA & {43.0} & {37.6} & {37.1} & {47.3} & {42.2} & {38.3} & {38.1} & {33.7} & {79.2} & {37.3} & {33.4} & {78.2} & {45.4}\\
\stddevs&&\withstd{0.0}&\withstd{3.5}&\withstd{2.4}&\withstd{2.0}&\withstd{3.1}&\withstd{4.5}&\withstd{1.4}&\withstd{1.1}&\withstd{1.6}&\withstd{1.5}&\withstd{1.4}&\withstd{2.8}&\withstd{0.8}\\
& CORAL & {40.3} & {38.7} & {38.3} & {47.9} & {40.3} & {40.2} & {38.2} & {33.8} & {81.7} & {38.8} & {35.0} & {84.0} & {46.4}\\
\stddevs&&\withstd{1.6}&\withstd{2.8}&\withstd{3.7}&\withstd{1.6}&\withstd{3.4}&\withstd{4.1}&\withstd{1.2}&\withstd{0.9}&\withstd{1.8}&\withstd{0.9}&\withstd{0.8}&\withstd{1.7}&\withstd{1.0}\\[\methodsep]
& RWA & {35.9} & {39.2} & {40.1} & {48.8} & {43.4} & {47.5} & {38.2} & {32.5} & {79.2} & {39.7} & {34.8} & {74.5} & {46.1}\\
\stddevs&&\withstd{2.4}&\withstd{4.4}&\withstd{4.5}&\withstd{3.7}&\withstd{4.3}&\withstd{6.1}&\withstd{2.1}&\withstd{2.2}&\withstd{3.2}&\withstd{1.6}&\withstd{1.1}&\withstd{3.9}&\withstd{1.3}\\
\hlinemid
\multicolumn{13}{l}{ResNet 50 features}\\[\subheadersep]
& Source SVM & {89.4} & {92.3} & {89.7} & {93.6} & {91.0} & {87.6} & {91.2} & {86.7} & {97.9} & {90.5} & {86.0} & {99.9} & {91.3}\\
\stddevs&&\withstd{1.2}&\withstd{2.6}&\withstd{2.5}&\withstd{0.7}&\withstd{2.3}&\withstd{3.2}&\withstd{1.0}&\withstd{1.0}&\withstd{1.2}&\withstd{1.0}&\withstd{0.8}&\withstd{0.2}&\withstd{0.7}\\[\methodsep]
& SA & {88.9} & {91.8} & {89.8} & {93.4} & {90.3} & {90.2} & {91.4} & {85.8} & {97.8} & {90.7} & {85.4} & {99.8} & {91.3}\\
\stddevs&&\withstd{1.3}&\withstd{2.7}&\withstd{1.3}&\withstd{0.7}&\withstd{2.2}&\withstd{2.1}&\withstd{1.0}&\withstd{0.9}&\withstd{1.1}&\withstd{1.0}&\withstd{1.0}&\withstd{0.4}&\withstd{0.5}\\
& CORAL & {89.2} & {92.2} & {91.9} & {94.1} & {92.0} & {92.1} & {94.3} & {87.7} & {98.0} & {92.8} & {86.7} & \best{100.0} & {92.6}\\
\stddevs&&\withstd{1.0}&\withstd{3.5}&\withstd{1.9}&\withstd{0.6}&\withstd{2.5}&\withstd{2.4}&\withstd{0.9}&\withstd{1.1}&\withstd{1.2}&\withstd{0.6}&\withstd{0.8}&\withstd{0.1}&\withstd{0.6}\\[\methodsep]
& RWA & \best{93.8} & \best{98.9} & \best{97.8} & \best{95.3} & \best{99.4} & \best{95.9} & \best{95.8} & \best{93.1} & \best{98.4} & \best{95.3} & \best{92.4} & {99.9} & \best{96.3}\\
\stddevs&&\withstd{0.7}&\withstd{1.7}&\withstd{2.0}&\withstd{0.5}&\withstd{0.7}&\withstd{3.1}&\withstd{0.1}&\withstd{0.5}&\withstd{1.3}&\withstd{0.5}&\withstd{0.5}&\withstd{0.2}&\withstd{0.4}\\
\hlinebot
\end{tabular}
\end{table*}

\begin{table}[t]
 \caption{Accuracy on the Office 31 dataset. }
 \label{tbl:accuracy-office-decaf7}
 \tablesize
 \centering
\begin{tabular}{@{\hspace*{\leftsep}}ll@{\hspace*{\colsep}}c@{\hspace*{\colsep}}c@{\hspace*{\colsep}}c@{\hspace*{\colsep}}c@{\hspace*{\colsep}}c@{\hspace*{\colsep}}c@{\hspace*{\colsep}}c}
\hlinetop
& & A$\to$D & A$\to$W & D$\to$A & D$\to$W & W$\to$A & W$\to$D & avg\\
\hlinemid
\multicolumn{7}{l}{DECAF-fc7 features}\\[\subheadersep]
& Source SVM & {58.4} & {53.1} & {43.2} & {86.3} & {43.6} & {90.4} & {62.5}\\[\methodsep]
& SA & - & - & - & - & - & - & -\\
& CORAL & {60.0} & {56.7} & {44.7} & {89.1} & {45.0} & {93.4} & {64.8}\\[\methodsep]
& RWA & {65.1} & {62.8} & {53.6} & {87.8} & {50.9} & {91.8} & {68.7}\\
\hlinemid
\multicolumn{7}{l}{ResNet features}\\[\subheadersep]
& Source SVM & {79.7} & {77.1} & {62.2} & {98.4} & {61.8} & \best{100.0} & {79.9}\\[\methodsep]
& SA & {79.9} & {79.1} & {63.2} & {98.1} & {61.5} & {99.8} & {80.3}\\
& CORAL & {80.9} & {77.6} & {58.8} & {98.6} & {59.9} & \best{100.0} & {79.3}\\[\methodsep]
& RWA & \best{90.0} & \best{90.6} & \best{74.4} & \best{99.0} & \best{73.7} & {99.6} & \best{87.9}\\
\hlinemid
\multicolumn{7}{l}{Deep Neural Networks (ResNet based)}\\[\subheadersep]
& DDC &  76.5  &  75.6    &62.2   & 96. 0   &61.5    &98.2   & 78.3 \\
 & DAN &   78.6  &  80.5   & 63.6   & 97.1  &  62.8    &99.6  &  80.4 \\
 & RTN &  77.5   & 84.5    &66.2   & 96.8    &64.8    &99.4    &81.6 \\
 & RevGrad & 79.7  &  82.0&   68.2  & 96.9   & 67.4    &99.1   & 82.2 \\
& JAN  & 84.7  &  85.4    &68.6       &97.4  &  70. 0   &99.8  &  84.3 \\
%
%
\hlinebot
\end{tabular}

\end{table}
%

%
%
  
\subsubsection*{Office dataset 31}
We next consider the standard Office dataset 31 \citep{saenko2010adapting} which contains 31 classes (the 10 from the Office-Caltech 10 plus 21 additional ones)  in 3 domains: Webcam (W), DSLR (D), and Amazon (A). 
In addition to Resnet, we also consider deep features from the 7th layer of AlexNet publicly available in \cite{tommasi2014testbed}, another deep neural network trained on Imagenet, which was also used in recent works, e.g. \cite{sun2016return}.
We found it beneficial to increase sparsity by rectifying these features, or equivalently, taking the activations of the networks after the rectifying nonlinearities. We then normalize by dividing by the standard deviation only.


%
We run experiments using all labeled source and unlabeled target data.
Results of experiments on this dataset are shown in \tblref{tbl:accuracy-office-decaf7}. These results show that {\cp } achieves state-of-the-art results, and outperforms other shallow DA algorithms by a large margin.  Also in this case, the gain in the accuracy is rather large when performing adaptation with  deep features over hard transfer tasks such as  {A$\to$W}  (from 77.1\% to 90.6\%) and on {D$\to$A} (from 62.2\% to 74.4\%). 

We also compare the results with deep features to those of other neural network based domain adaptation methods.
{\cp }  achieves results comparable or better than those obtained by  end-to-end methods based on deep neural networks.  Execution of SA with DECAF features did not terminate after 5 days.
Note that the considered deep end-to-end methods use  weights pre-trained on Imagenet. Furthermore  RTN uses target labels to perform model selection. 

 

\subsubsection*{Cross Dataset Testbed}
Finally we consider a larger scale evaluation using the Cross Dataset Testbed \citep{tommasi2014testbed}, again using rectified deep features obtained with DECAF. The dataset contains $40$ classes from 3 domains: $3847$ images for the domain Caltech256 (C), $4000$ images for Imagenet (I), and $2626$ images for SUN (S).  Results of these experiments are shown in \tblref{tbl:accuracy-cdt}.
Also on this dataset {\cp} obtains best results, and improves by a large margin over no adaptation. 
Previous papers have used standardization of the features.

\begin{table}[t]
 \caption{Accuracy on the Cross Dataset Testbed.}
 \label{tbl:accuracy-cdt}
 \tablesize
 \centering
\begin{tabular}{@{\hspace*{\leftsep}}ll@{\hspace*{\colsep}}c@{\hspace*{\colsep}}c@{\hspace*{\colsep}}c@{\hspace*{\colsep}}c@{\hspace*{\colsep}}c@{\hspace*{\colsep}}c@{\hspace*{\colsep}}c}
\hlinetop
& & C$\to$I & C$\to$S & I$\to$C & I$\to$S & S$\to$C & S$\to$I & avg\\
\hlinemid
& Source SVM & {68.7} & {22.4} & {76.2} & {24.9} & {29.5} & {30.5} & {42.0}\\[\methodsep]
& SA & {68.8} & {23.0} & {74.9} & {24.9} & {30.5} & {31.1} & {42.2}\\
& CORAL & {69.0} & {23.6} & {75.9} & {25.7} & {34.8} & {34.2} & {43.9}\\[\methodsep]
& RWA & \best{74.5} & \best{25.1} & \best{80.5} & \best{27.4} & \best{40.9} & \best{42.8} & \best{48.5}\\
\hlinebot
\end{tabular}
\end{table}

\pgfplotscreateplotcyclelist{oversamples}{%
  red,            mark repeat=50, every mark/.append style={fill=.!80!red},  mark=*\\%
  blue,           mark repeat=50, every mark/.append style={fill=.!80!blue}, mark=square*\\%
  green!60!black, mark repeat=50, every mark/.append style={fill=.!80!black},mark=triangle*\\%
  black,          mark repeat=50, mark=star\\%
}
\pgfplotscreateplotcyclelist{oversamples-zoom}{%
  red,            mark repeat=5, every mark/.append style={fill=.!80!red},  mark=*\\%
  blue,           mark repeat=5, every mark/.append style={fill=.!80!blue}, mark=square*\\%
  green!60!black, mark repeat=5, every mark/.append style={fill=.!80!black},mark=triangle*\\%
  black,          mark repeat=5, mark=star\\%
}
\begin{figure}[ht]
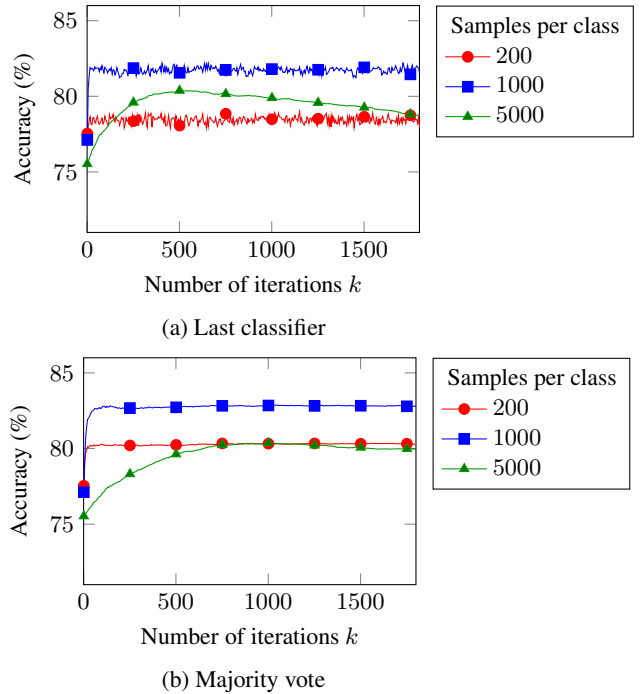

  \centering
  \begin{subfigure}[b]{1\linewidth}
    \centering

    \caption{Last classifier\hspace*{20mm}}
  \end{subfigure}\\[.9ex]
  \begin{subfigure}[b]{1\linewidth}
    \centering
    %
    \label{fig:num-iterations-ensemble}
   \caption{Majority vote\hspace*{20mm}}
  \end{subfigure}
  \caption{Average accuracy of {\cp} over all domains of the Amazon dataset, as a function of the number of iterations and the number of bootstrap samples per class (the dataset has 1000 samples per class).}
  \label{fig:num-iterations}
\end{figure}

\subsection{Sensitivity analysis}\label{sec:sensitivity}

{\cp} is a stochastic method, since it relies on bootstrap samples in each iterations.
Experiments indicate that the variance in the results over multiple runs is small.  For instance, average standard deviation of the accuracy over 10 repetitions on the full Office-Caltech dataset with
SURF features is $0.9$, and on the on the Amazon dataset it is $0.1$, which shows that the method is robust.

 {\cp}  has two parameters: $K$, the number of iterations, and $m$,  the size of the bootstrap sample. We investigated empirically how the choice of these parameters influences the results. \Figref{fig:num-iterations} shows the results of these experiments on the Amazon sentiment analysis dataset.
We can see that if the bootstrap sample is too large, then the iterates $h^k$ are very similar, and the method can take a long time to converge.
But on the other hand, if the sample is too small, the iterates become noisy, and have a low accuracy. By itself, noisy iterates are not a problem, since the ensemble prediction will still be good, as can be seen in \figref{fig:num-iterations}b. Although  our sensitivity analysis  indicates a good convergence behaviour of  {\cp }, in general {\cp}  is not guaranteed to converge, as shown for instance by our toy example. 

\section{Conclusions}
 
We presented a new method for unsupervised DA based on a random sampling strategy controlled by a given source hypothesis. 

Our final majority vote classifier is piecewise linear.  This may be a limitation in case of highly non-separable domains. All other baselines here considered use more involved forms of non-linearity, either in the features they construct, or in the architecture. 
Although results of our experiments showed that on visual domain adaptation tasks  {\cp} profits from the use of deep features from pre-trained models and performs on par with end-to-end deep neural networks methods, it remains to be investigated whether non-linear extensions of {\cp} could be even more effective.

When the source classifier assigns the same label to all target instances, the corresponding state in our Markov chain has a self-loop with transition probability $1$. By assuming irreducibility of the Markov chain this case is ruled out.  A less strong assumption is the reachability of the true target labeling from the initial labeling based on $h_s$. It remains to be investigated how this desirable property can be characterized or enforced in terms of domain discrepancy.

{\cp } combines source and target hypotheses by their average.  More sophisticated integration techniques, like hypothesis transfer, e.g. \cite{scholkopf2001generalized}, could possibly be more beneficial than our simple choice.




\bibliography{da}

\bibliographystyle{named}

\vfill

\end{document}